\documentclass[10pt,twocolumn,letterpaper]{article}

\usepackage[pagenumbers]{cvpr} % To force page numbers, e.g. for an arXiv version

\usepackage{tikz}
\usepackage{booktabs}
\usepackage{multirow}
\usepackage{multicol}
\usepackage{colortbl}
\usepackage{xcolor}
\usepackage{graphicx} 
\usepackage{array}
\usepackage{xcolor}
\usepackage{graphicx} 
\usepackage{arydshln}
\usepackage{float}

\newcommand{\ours}{EVATok}
\makeatletter
\def\blfootnote{\xdef\@thefnmark{}\@footnotetext}
\makeatother

\definecolor{cvprblue}{rgb}{0.21,0.49,0.74}
\usepackage[pagebackref,breaklinks,colorlinks,allcolors=cvprblue]{hyperref}

\def\paperID{3636} % *** Enter the Paper ID here
\def\confName{CVPR}
\def\confYear{2026}

\title{EVATok: Adaptive Length Video Tokenization for Efficient Visual Autoregressive Generation}

\author{
    Tianwei Xiong$^1$ \quad
    Jun Hao Liew$^2$ \quad
    Zilong Huang$^2$ \quad
    Zhijie Lin$^2$ \quad
    Jiashi Feng$^2$ \quad
    Xihui Liu$^{1\dagger}$ \\[6pt]
    $^1$The University of Hong Kong \qquad
    $^2$ByteDance Seed\\
}

\begin{document}
\maketitle
% \twocolumn[{
% \maketitle
%    \centering
%     \includegraphics[width=\linewidth]{figures/teaser_2col.pdf}
%     \captionsetup{type=figure}
%     \vspace{-0.7cm}
%     \captionof{figure}{%
%     \textbf{Teaser.} 
%     }
%     \label{fig:teaser}
% }]

\blfootnote{Project page: \href{https://silentview.github.io/EVATok/}{https://silentview.github.io/EVATok/}}
\begin{abstract}
Autoregressive (AR) video generative models rely on video tokenizers that compress pixels into discrete token sequences. The length of these token sequences is crucial for balancing reconstruction quality against downstream generation computational cost. Traditional video tokenizers apply a uniform token assignment across temporal blocks of different videos, often wasting tokens on simple, static, or repetitive segments while underserving dynamic or complex ones. 
To address this inefficiency, we introduce \textbf{EVATok}, a framework to produce \textbf{E}fficient \textbf{V}ideo \textbf{A}daptive  \textbf{Tok}enizers. Our framework estimates optimal token assignments for each video to achieve the best quality-cost trade-off, develops lightweight routers for fast prediction of these optimal assignments, and trains adaptive tokenizers that encode videos based on the assignments predicted by routers. 
We demonstrate that EVATok delivers substantial improvements in efficiency and overall quality for video reconstruction and downstream AR generation. Enhanced by our advanced training recipe that integrates video semantic encoders, EVATok achieves superior reconstruction and state-of-the-art class-to-video generation on UCF-101, with at least 24.4\% savings in average token usage compared to the prior state-of-the-art LARP and our fixed-length baseline. 

\end{abstract}
\begin{figure}[t!]
% \vspace{-0.1in}
    \centering
\includegraphics[width=\linewidth]{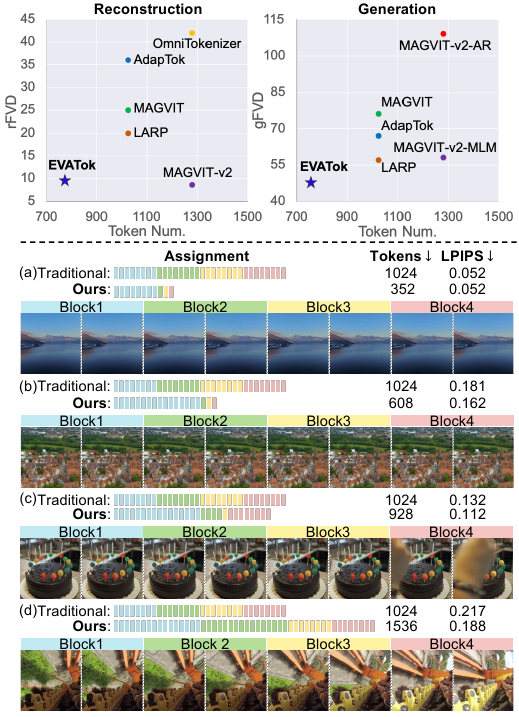}
    \vspace{-0.26in}
    \caption{
    \textbf{\ours{} highlights.} 
    \textbf{Top}: \ours{} achieves superior video reconstruction and downstream generation quality with significant savings in token usage.  
    \textbf{Bottom}: \ours{} assigns tokens in an intuitive way. Clips with dynamic motion or complex layout will be encoded with more tokens, while clips that are repetitive or simple will be assigned fewer tokens.
    % \ours{} assign tokens adaptively to different temporal blocks across videos. Simple-layout and repetitive blocks are assigned limited token budgets, while blocks with a complex layout and dynamic content receive more tokens.
    % our content-adaptive video tokenization and adaptive length generation workflow. Videos with complex layout and dynamic motion will be assigned more tokens, while static or simple-layout videos will be assigned fewer. 
    % \tianwei{Messy. Use bars. Pick single object moving fast. More obvious. 3 can be more complex. Can illustrate. 
    }
    \label{fig:teaser}
    \vspace{-0.2in}
\end{figure}

\section{Introduction}
\label{sec:intro}

% background: AR (LLM, vision); tokenizers; length and quality.
Visual generation with autoregressive~(AR) language models is rapidly advancing~\cite{lumina, wang2024emu3, cui2025emu35, xin2025lumina-mgpt2.0}, driven by the success of LLMs~\cite{llama3, comanici2025gemini2.5, gpt4, deepseek-r1, deepseek-v3} and their potential for unified multi-modal generation~\cite{videopoet, wu2024vila, qu2024tokenflow, janus-pro, liquid}. AR visual generative models typically operate on sequences of discrete visual tokens, obtained by patch-wise compression of pixels via visual tokenizers~\cite{vqgan, vit_vqgan}. The tokenizer’s design directly influences reconstruction quality and token sequence length, thus determining the quality-cost trade-off and computational overhead for downstream AR models.

% governs both the reconstruction quality and the token sequence length, thereby shaping the quality-cost trade-off and computational overhead for downstream AR models.

% problem: why assigning the same budget to different samples is suboptimal. And why is uniform assignment for different temporal blocks to videos suboptimal?
However, most existing visual tokenizers~\cite{vqgan, vit_vqgan, magvit-v2} produce fixed-length sequences regardless of input content complexity. This uniform budget allocation is especially inefficient for AR video generative models using causal video tokenizers~\cite{wang2024emu3, wang2024loong, videopoet, agarwal2025cosmos, magvit-v2}, as information density in videos varies not only across samples but also temporally: simple-layout, near-static or repetitive segments receive excessive tokens, while dynamic or complex-layout segments are undeserved, compromising both efficiency and fidelity.

% previous solution, what is desirable but missing
Ideally, given a video and specified preference between better quality or less token cost, we would predict an optimal \textit{assignment}---specifying the total number of tokens used for video reconstruction and their distribution over temporal blocks---that maximizes the quality-cost trade-off. Prior video adaptive tokenizers~\cite{yan2025elastictok, Li2025AdapTok} enable variable length compression via tail-token-dropping~\cite{Miwa2025OneDPiece, flextok} training, with assignment selection by threshold-based search~\cite{yan2025elastictok} or Integer Linear Programming~(ILP) within video mini-batches under fixed average budget constraints~\cite{Li2025AdapTok}. 
However, these approaches can yield suboptimal results: heuristic threshold-based searches may neglect global quality-cost balance, while mini-batch ILP ties per-sample decisions to the batch compositions and rigid average budgets. Critically, they do not address the core need: for each sample, determining the optimal assignment tailored to the samples's inherent complexity, enabling optimal adaptive tokenization that allocates budgets where they are most needed, achieving the best balance for overall efficiency and quality. 

% optimal assignment estimation
The challenge for optimal assignment identification is that there was no estimation approach or even definition for it. To fill in this blank, we formulate the optimal assignment identification problem as a tractable maximum proxy reward assignment identification task, where the proxy reward is a novel metric measuring both the reconstruction quality and cost (token length) to quantify the quality-cost trade-off for a particular assignment. In other words, the optimal assignment with the maximum proxy reward achieves the best quality-cost trade-off.

% that we introduce to quantify the quality-cost trade-off performance for a particular assignment. The optimal assignment with the maximum proxy reward achieves the best quality-cost trade-off. For a video, the proxy reward of an assignment is an aggregated value combining the weighted quality and cost, where the quality is measured by reconstruction with a variable length tokenizer and the cost is the total token length.
% to estimate proxy reward
To estimate the proxy reward, we introduce a proxy tokenizer that learns to reconstruct the input video under different token assignments. Once trained, we can simply iterate over all possible candidates to identify the optimal token assignment with maximum proxy reward.
% By brute-force searching for maximum proxy reward assignment for each video, we implement the optimal assignment identification. 
And to build a faster approach for optimal assignments prediction, we curate a dataset to train a lightweight model, named the router, which learns optimal assignment prediction in a classification task form. 
% the application of the router
Equipped with this router, we train final adaptive tokenizers to encode videos using content-adaptive assignments, which in turn support downstream efficient adaptive length AR generative models. 
In summary, \ours{} unfolds in four stages:~(1) Train a proxy tokenizer for optimal assignment estimation;~(2) Curate a dataset of ~(video, optimal assignment) pairs for router training;~(3) Train a lightweight router for fast optimal assignment prediction; and~(4) Train the final video adaptive tokenizer under assignments from the router.

For video reconstruction and downstream AR generation, \ours{} yields substantial gains in efficiency and quality. Enhanced by our advanced recipe integrating video semantic encoders~\cite{tong2022videomae, vjepa2} into tokenizer training, \ours{} achieves superior reconstruction and state-of-the-art~(SOTA) class-to-video generation quality with at least 24.4\% token length savings compared to prior video tokenizers~\cite{larp, Li2025AdapTok}, as shown in Fig.~\ref{fig:teaser}. The adaptive assignment examples of \ours{} in Fig.~\ref{fig:teaser} also correspond to intuitions: repetitive, simple-layout, and static content is assigned fewer tokens; in contrast, non-repetitive, complex-layout, and dynamic content is assigned more. \ours{} highlights the promising potential of content-adaptive video tokenization for improving efficiency and quality for overall reconstruction and downstream AR generation.

% summary for contributions
We summarize our main contributions as follows:
\begin{itemize}
  \item A four-stage framework for efficient video adaptive tokenization, featuring a router that provides optimal budget assignment during training and inference of tokenizers.
   \item Proxy reward: a novel metric utilizing a variable length tokenizer to identify optimal assignments for each video.
   % to provide supervision for adaptive assignment task: a novel, practical mechanism to learn per-sample optimal assignments under explicit quality–cost preferences using a proxy tokenizer.
  % \item Proxy reward to provide supervision for adaptive assignment task: a novel, practical mechanism to learn per-sample optimal assignments under explicit quality–cost preferences using a proxy tokenizer.
  \item Extensive experiments showing that content-adaptive video tokenization can surpass fixed-length baselines, achieving superior performances in reconstruction and downstream AR generation with fewer tokens.
\end{itemize}

\section{Related Work}
\label{sec:related work}

\noindent\textbf{Discrete image and video tokenizers.} 
Since the classic VQ-VAE~\cite{vqvae} and VQ-GAN~\cite{vqgan}, extensive efforts have been made to better compress visual inputs into discrete token sequences for autoregressive modeling. LFQ~\cite{magvit-v2} and FSQ~\cite{FSQ} are proposed for large-scale codebook training. VAR~\cite{var} encodes token sequences in a residual-style~\cite{rq} multi-scale structure for efficient generation. For videos, while many works choose 3D CNN to implement video tokenizers~\cite{ge2022TATS, magvit-v2, wang2024emu3, wang2024loong}, recently more video tokenizers are being implemented using transformer architecture~\cite{wang2024omnitokenizer, larp, villegas2022phenaki, yan2025elastictok, Li2025AdapTok}. Transformers are beneficial to video tokenizers not only due to their known scalability, but also because their flexible attention mechanism naturally helps build 1D tokenizers~\cite{yan2025elastictok, titok}, which removes the grid-like spatial prior in token sequences, making the sequence length easy to adjust and convenient for adaptive tokenization. In this work, we use Q-Former-style~\cite{DETR, blip2} 1D tokenizer design~\cite{Xiong2025GigaTok} to build adaptive video tokenizers.

% C-ViViT, magvit-v2, OmniTokenizer, LARP...

\noindent\textbf{Adaptive visual tokenization.} 
Based on the intuition that simple content needs fewer tokens while complex content needs more for efficient compression, the seminal work Dynamic VQ~\cite{dynamicvq} encodes different regions across images with different granularity adaptively, utilizing Gumbel Softmax~\cite{gumbel-softmax}. Differently, CAT~\cite{Shen2025CAT} lets LLMs decide the compression granularity based on captions. Recently, techniques like tail-token-dropping~\cite{Miwa2025OneDPiece, flextok, Wang2025ALTo, yan2025elastictok} or iterative token allocation~\cite{duggal2025ALIT, mao2025dove} have been used to enable tokenizers to encode visual inputs under different token assignments. Further on video tokenization, ElasticTok~\cite{yan2025elastictok} and AdapTok~\cite{Li2025AdapTok} study how to determine these given assignments in adaptive video tokenization. However, their adaptive assignment searching methods are heuristic and can potentially lead to suboptimal assignments. A concurrent work, InfoTok~\cite{ye2026infotok}, masks less important tokens from pre-trained tokenizers with an ELBO-based method.
In this work, \ours{} directly predicts the optimal assignments given input videos and preferences between quality and cost.

\noindent\textbf{Video representation alignment.} The representation of pretrained semantic encoders~\cite{dinov2, CLIP, siglip} have been used to enhance image generative models~\cite{repa} or  image tokenizers~\cite{repa_in_vae, Xiong2025GigaTok, lightningDiT, factorized_vq_repa, ma2025unitok}. Recently, similar approaches have been studied for video diffusion models~\cite{zhang2025videorepa} or reported in use for video tokenizers~\cite{cui2025emu35}. We further reveal that representation alignment is beneficial for video tokenizers, especially when combined with semantic video discriminators.

% for video tokenizers, representation alignment is beneficial, especially when combined with semantic video discriminators.

% \noindent\textbf{Autoregressive visual generation.} 
% Extensive prior works have explored the scalable training of AR models in image and video generation~\cite{llamagen, lumina, xin2025lumina-mgpt2.0, liquid, videopoet, wang2024loong}. And targeting better AR visual generation, significant efforts~\cite{Xiong2025GigaTok, magvit-v2, openmagvit-v2, ibq_scale, ma2025unitok} have been put into improving image tokenizers. 
% In this work, besides introducing better tokenizer training recipe for AR video generation, we also reveal how adaptive tokenizers can improve the quality-cost trade-off for AR visual generation.

% directly as image tokenizer encoders~\cite{zheng2025rae}. Recently, the semantic video encoders are also reported to benefit video generation models~\cite{zhang2025videorepa}. In this work, we further reveal for video tokenizers, representation alignment is beneficial especially when combined with pre-trained semantic video discriminators.

% Image: Lookup ViT, Dynamic VQ, One-D-Peace, ALIT, DOVE, CAT, FlexTok
% Video: ElasticTok, AdapTok
% Others: ALTo

% REPA, REPA-E, RAE, VideoREPA
% GigaTok, VA-VAE, UniTok, VILA-U, ...

\section{Method}
\label{sec:methods}
\begin{figure*}[t]
% \vspace{-0.1in}
    \centering
    \includegraphics[width=\linewidth]{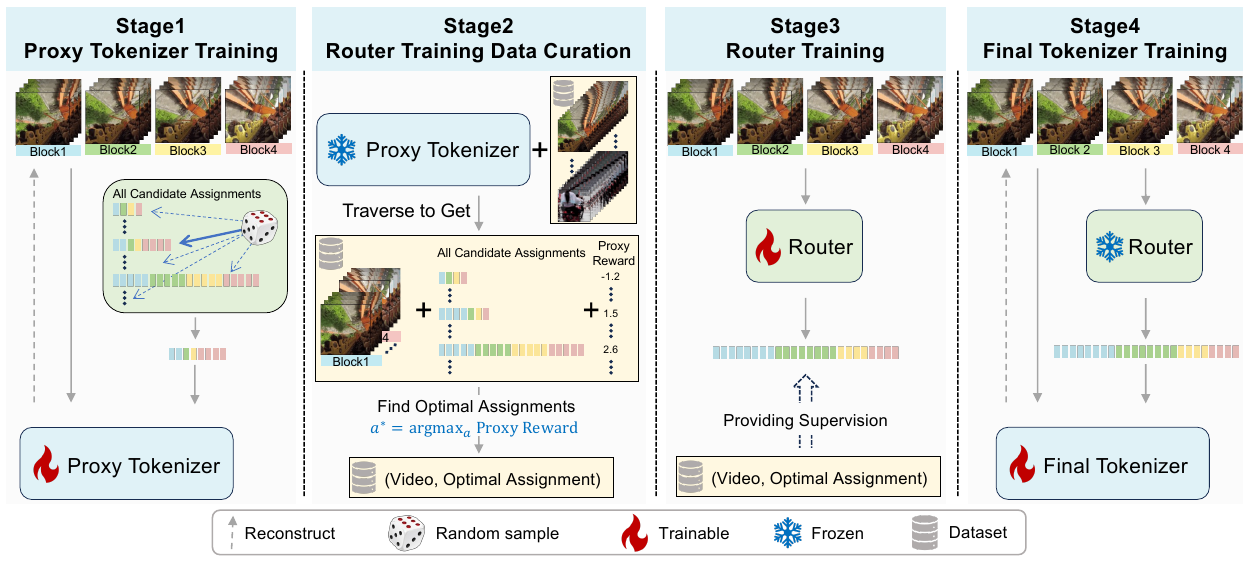}
    % \vspace{-0.30in}
    \caption{\textbf{Four-stage framework for adaptive video tokenizer training.} \textbf{Stage 1} trains a proxy tokenizer to reconstruct videos under all candidate assignments. \textbf{Stage 2} applies the proxy tokenizer to compute proxy rewards for all candidate assignments across videos from a dataset. It identifies the assignments with maximum proxy rewards to curate a classification dataset of videos and their optimal assignments. 
    \textbf{Stage 3} trains a router on the curated dataset to predict the optimal assignments for videos.
    \textbf{Stage 4} trains the final tokenizer from scratch, with the router determining the assignment for each input video during training.}
    % \caption{\textbf{Three-stage framework for adaptive video tokenizer training.} \textit{Stage 1}: a proxy tokenizer is trained to reconstruct videos with all candidate assignments. \textit{Stage 2}: the proxy tokenizer is used to calculate the proxy reward of all candidate assignments for all videos from a dataset~(in practice, subset of WebVid-10M~\cite{Bain21WebVid}). Then, by finding the max-proxy-reward assignments, a classification dataset can be curated and used to train router. \textit{Stage 3}: the final tokenizer is trained from scratch, and the ViT router will decide the assignment for each video input during training.}
    \label{fig:framework}
        \vspace{-0.10in}

\end{figure*}

\medskip \noindent \textbf{Problem setup.}
We first introduce the problem setup of our video adaptive tokenizer before presenting our proposed solution. 
Given a video $x$, we temporally downsample it by $4\times$ and divide it into $T$ causal blocks. Each block $t$ is assigned $k_t$ tokens from a candidate set $K$ (\eg, $\{32, \dots, 512\}$) with $m$ levels, forming an assignment $a = (k_1, \dots, k_T)$ with total token length $L(a) = \sum_{t=1}^T k_t$. We identify that the primary challenge for an adaptive video tokenizer lies in predicting the optimal token assignment $a^*$ for each video to achieve the best quality-cost trade-off.

We formulate the optimal assignment prediction problem as a tractable maximum proxy reward assignment prediction task, where the proxy reward is a novel metric that we introduce to quantify the quality-cost trade-off performance for a particular token assignment. Centering on the idea of using proxy reward for optimal assignment prediction, we build our four-stage framework, as in Fig.~\ref{fig:framework}, for efficient video adaptive tokenization:
\textbf{~(Stage 1)} train a proxy tokenizer that can reconstruct videos \wrt randomly sampled token assignments. This proxy tokenizer later serves for proxy reward computation; \textbf{~(Stage 2)} curate a dataset comprising~(video, optimal token assignment) pairs by searching proxy reward under different token assignments calculated with the proxy tokenizer. This dataset serves for training a router for fast optimal assignment prediction; \textbf{~(Stage 3)} train the router on this curated dataset, to accelerate optimal assignment prediction largely against searching; \textbf{~(Stage 4)} train an adaptive video tokenizer using the optimal assignments predicted by the router, hence achieving adaptive length video tokenization. 
Next, we will introduce each stage with more details.

\subsection{Stage 1: Training a Proxy Tokenizer}
\begin{figure}[t]
% \vspace{-0.1in}
    \centering
    \includegraphics[width=\linewidth]{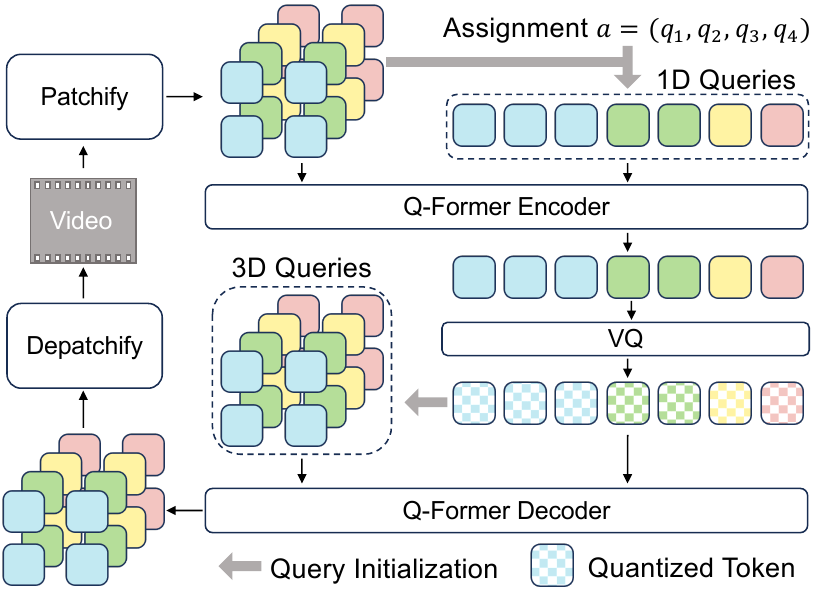}
    % \vspace{-0.30in}
    \caption{\textbf{Architecture of 1D variable-length video tokenizer for \ours{}.} The input video is spatio-temporally patchified into 3D embeddings. According to a given assignment $a$, 1D variable-length query embeddings are initialized from these 3D embeddings. After Q-Former encoding and quantization, 1D discrete tokens are produced. Finally, 3D queries are initialized to reconstruct the video frames from the 1D tokens.} 
    % \caption{\textbf{Architecture of our 1D variable-length video tokenizer.} Input video will be spatial-temporally patchified into 3D embeddings (different colors represent different time blocks), and according to a given assignment $a$, 1D query embeddings will be initialized from the 3D embeddings, (different temporal blocks can have different query lengths). After Q-Former encoding and quantization, 1D discre tokens are produced and 3D queries will be initialized and reconstruct the video from the 1D tokens.
    % }
    \label{fig:tokenizer_1d}   
    \vspace{-0.20in}

\end{figure}
In stage 1, we train a proxy tokenizer that can reconstruct a video \wrt a randomly sampled assignment $a$. This proxy tokenizer can serve as a proxy for the assessment of the quality of a particular token assignment, which we subsequently use to identify optimal assignments, train a router, and build the final adaptive length tokenizer.

\medskip \noindent \textbf{Model architecture.} We adopt a Q-Former~\cite{blip2, DETR} style 1D tokenizer for its scalability~\cite{Xiong2025GigaTok} and flexibility in variable length tokenization. As shown in Fig.~\ref{fig:tokenizer_1d}, input videos are first spatio-temporally patchified into 3D embeddings using a simple linear patch embedding module, consistent with prior video tokenizers~\cite{larp, Li2025AdapTok, yan2025elastictok}. Then, given a randomly sampled token assignment $a=(q_1, q_2, q_3, q_4)$ specifying the number of 1D tokens per temporal block, a 1D query sequence is initialized, \ie, each 1D query is derived from the 2D pooling feature of the corresponding temporal block in the 3D embeddings. Through Q-Former encoder layers, these 1D queries encode visual information from the 3D embeddings and are vector-quantized into discrete tokens. Temporally causal attention masks ensure that 1D tokens do not encode information from subsequent blocks. 

As for the decoder, 3D queries are initialized using the first 1D token in their corresponding temporal block. After a similar temporally causal decoding process, the final 3D features will be linearly projected and reshaped into video frames. We do not use the typical tail-token-dropping~\cite{Miwa2025OneDPiece, yan2025elastictok, Li2025AdapTok, mao2025dove} strategy to produce variable length tokens as it may lead to two concerns:~(1) extra computation overhead caused by using many tokens that will just be dropped as register tokens;~(2) the roles of tail 1D queries being ambiguous during encoding: tail 1D queries cannot be aware of whether they will be dropped after encoding. Since the two concerns potentially hurt efficiency and performance, in \ours{}, the length of 1D tokens is decided and fixed during the initialization of 1D queries. 

\medskip \noindent\textbf{Training recipe.} We enhance the tokenizer training by video semantic encoders~\cite{tong2022videomae, vjepa2} through video representation alignment. Following typical image representation alignment approaches~\cite{repa, lightningDiT, Xiong2025GigaTok}, we apply patch-wise alignment between the intermediate 3D features of the tokenizer decoder and the features from pre-trained V-JEPA2-L~\cite{vjepa2}. We use a linear projection and reshape strategy, similar to depatchify, to address feature shape mismatches. Formally, let $f^{\text{dec}, l}$ be the output 3D features from the $l$-th decoder layer, and $f^{\rm sem}$ the semantic features from the pretrained encoder. The representation alignment loss is:
\begin{equation}
\label{eq:sem reg}
    \mathcal{L}_{\rm align} = -\frac{1}{N} \sum_{n=1}^N  {\rm sim} \Big( f^{\text{dec},l}_n, \phi(f^{\rm sem}_n) \Big)
\end{equation}
where $N$ is the batch size, $n$ is the batch item index, $\text{sim}(\cdot, \cdot)$ is cosine similarity, and $\phi(\cdot)$ combines an MLP and a depatchify module for shape matching. 
We use a transformer PatchGAN~\cite{patchgan} discriminator as in LARP~\cite{larp}. The final training loss of our video tokenizer is:
\begin{equation}\label{eq:loss}
    \mathcal{L}_{\text{total}} = \mathcal{L}_{\text{vqgan}} + \lambda \mathcal{L}_{\text{align}} +  
    \gamma \mathcal{L}_{\text{entropy}}
\end{equation}
with $\lambda$ tuned as $0.7$ by default. Here, $\mathcal{L}_{\text{vqgan}}$ combines $l_1$ reconstruction loss $\mathcal{L}_{\text{recon}}$, perceptual loss $\mathcal{L}_{\text{percp}}$~\cite{LPIPS, johnson2016perceptual}, adversarial loss $\mathcal{L}_{\text{GAN}}$, and VQ codebook loss $\mathcal{L}_{\text{VQ}}$~\cite{vqgan, vit_vqgan}; $\mathcal{L}_{\text{entropy}}$ is the entropy loss from~\cite{magvit-v2} used for better codebook usage, with $\gamma$ set empirically as $0.02$.
% Here $\mathcal{L}_{\text{vqgan}}$ is a combination of multiple losses, including $\mathcal{L}_{\text{recon}}$, the $l_1$ reconstruction loss on video pixels, $\mathcal{L}_{\text{percp}}$, the perceptual loss~\cite{LPIPS, johnson2016perceptual}, $\mathcal{L}_{\text{GAN}}$, the adversarial loss, and $\mathcal{L}_{\text{VQ}}$~\cite{vqgan, vit_vqgan} the VQ codebook loss. $\mathcal{L}_{\text{entropy}}$ is entropy loss~\cite{magvit-v2} and $\gamma$ is empirically set as $0.02$.

\subsection{Stage 2: Dataset Curation for Router Training}
% Finding the optimal assignment requires a measurement for the quality-cost trade-off performance of an assignment $a$ for video $x$. 
With the proxy tokenizer, we can use it to assess the quality-cost trade-off performance of a specific assignment $a$ for a video $x$, which will be illustrated in detail later. This means we can brute-force evaluate all the candidate assignments for $x$ to find the optimal one. However, brute-force searching is undesirable due to the massive computational cost during adaptive video tokenization. Therefore, we aim to train a lightweight router that predicts optimal assignments in one pass. Towards this objective, we design stage 2 to curate a dataset to train such a lightweight router.

First, we illustrate how to evaluate the quality-cost trade-off performance of $a$ on $x$ with the proxy tokenizer. We quantify this performance using the \textbf{proxy reward}:
\begin{equation}
\label{eq:proxy reward}
    R_\text{proxy} = w_q Q(\mathcal{E}_\text{proxy}, x, a) - w_l L(a) 
\end{equation}
where $Q(x, a)$ denotes the reconstruction quality of $a$ for $x$, $L(a)$ is the token length cost of $a$, and $w_q, w_l$ are the weights reflecting preferences for better quality or less cost. For each $x$, its optimal assignment $a^*$ maximizes $R$, balancing token savings with minimal quality loss. Then, with this measurement, we resolve the challenging task of optimal assignment $a^*$ prediction by brute-force searching for the $a$ with maximum proxy reward:
\begin{equation}
a^* = \text{argmax}_{a\in A}R_\text{proxy}
\end{equation}

% , we can search for the optimal assignment maximizing the quality-cost trade-off for reconstruction. In this work, we formulate the measurement as a proxy reward combining the quality and cost:
% \begin{equation}
% \label{eq:proxy reward}
%     R_\text{proxy} = w_q Q(\mathcal{E}_\text{proxy}, x, a) - w_l L(a) 
% \end{equation}
% where $Q(x, a)$ denotes the reconstruction quality of $a$ for $x$, $L(a)$ is the token length cost of $a$, and $w_q, w_l$ are the weights reflecting preferences for better quality or less cost. For each $x$, its optimal assignment $a^*$ maximizes $R$, balancing token savings with minimal quality loss. 
% With this measurement, we resolve the challenging task of optimal assignment $a^*$ prediction by searching for the $a$ with maximum proxy reward:
% \begin{equation}
% a^* = \text{argmax}_{a\in A}R_\text{proxy}
% \end{equation}

In practice, we collect 100k video clips from the diverse dataset WebVid-10M~\cite{Bain21WebVid}. We record the reconstruction quality~(LPIPS~\cite{LPIPS}, PSNR, MSE) of each video clip under all candidate assignments. Then, with specified preference weights $w_q, w_l$, the proxy reward is calculated with Eq.~\ref{eq:proxy reward} for all candidate assignments for each video. Specifically, we calculate $Q(\mathcal{E}_\text{proxy}, x, a)$ as the normalized LPIPS~\cite{LPIPS}, and $L(a)$ as the normalized length. Finally, only the assignment with the maximum proxy reward will be chosen for the video, resulting in a training dataset of 100k videos and their respective ground-truth assignments.

% calculated for all candidate assignments for each video. Only the assignment with the maximum proxy reward will be chosen for the video, resulting in a training dataset of 100k videos and their respective groundtruth assignments.

\subsection{Stage 3: Router Training}
With the training dataset containing~(video, optimal assignment) pairs from stage 2, we can train a lightweight router for one-pass optimal token assignment prediction in a classification formulation task. Our router adopts a ViT-like architecture~\cite{vit} and is trained to classify input video $x$ into one of the $m^T$ candidate assignment categories, which should be the optimal assignment for $x$.
Given an input video, our router patchifies it to 3D visual embeddings, appends a {\tt [CLS]} embedding. The router finally produces the probability of each candidate assignment $a$ being the optimal one for the input video from the {\tt [CLS]} embedding feature and is trained with cross-entropy loss.

% The router is trained with cross-entropy loss, similar to how ViT is trained in the image classification task~\cite{russakovsky2015imagenet, vit}.

% Searching for the optimal assignment by calculating the proxy reward for all candidate assignments leads to a heavy inference-time computational cost for adaptive tokenization. To mitigate this issue, we train a lightweight router for one-pass optimal token assignment prediction using the curated training data from stage 2. 

% predicts the probability of each candidate assignment $a \in A$ being the optimal one with the highest proxy reward 

% is trained using cross-entropy loss, and its predicted probability of each candidate assignment $a \in A$ being the optimal one with the highest proxy reward

% the {\tt [CLS]} embedding will eventually be processed to the probability of each candidate assignment $a \in A$ being the optimal one with the highest proxy reward. 

% \junhao{slightly extend this? For example, the loss, and prediction head \etc}

\subsection{Stage 4: Adaptive Length Video Tokenizer}
Integrating the router into our final video adaptive tokenization solution, we train an adaptive length video tokenizer conditioned on the token assignments predicted by the router. Specifically, the router predicts the optimal assignment for each video clip sample, which decides the token length and temporal distribution of the encoded token sequence. And the adaptive tokenizer learns to reconstruct each video using the predicted assignment from the router. Instead of combining the router with the proxy tokenizer as the final video adaptive tokenization solution, we choose to train a final tokenizer from scratch with the assignments from the tokenizer. This is to mitigate an issue in proxy tokenizer and prior video adaptive length tokenizers~\cite{Li2025AdapTok, yan2025elastictok}: the training-inference gap. For the proxy tokenizer, it is trained to encode videos across all $m^T$ possible assignments, yet during inference, only a few assignments might be used per video. This inefficiency in training can degrade tokenizer performance, as shown in Sec.~\ref{sec: final validation}, and is addressed by \ours{} in our stage 4 training.

Different from the proxy tokenizer, which can suffice with a simpler training recipe, the final tokenizer training can further benefit from advanced training designs. Inspired by DINO~\cite{dino, dinov2} discriminators in image tokenization~\cite {var, ma2025unitok}, in the final tokenizer training, we optionally employ video semantic encoders as discriminators for potentially better reconstruction and downstream AR generation.  We use a frozen pretrained VideoMAE-B~\cite{tong2022videomae, wang2022internvideo} to process input videos and feed multi-layer features to trainable 1D CNN heads for fake/real logits. We avoid larger V-JEPA2 models~\cite{vjepa2} due to logit divergence instability risks~\cite{lu2025atoken} in adversarial training.
We validate that a VideoMAE semantic discriminator, combined with video representation alignment, can significantly enhance both reconstruction and downstream AR generation quality for video tokenizers in Sec.~\ref{sec:ablation}.

\section{Experiments}
\label{sec:experiments}
\begin{figure*}[t]
% \vspace{-0.1in}
    \centering
    \includegraphics[width=\linewidth]{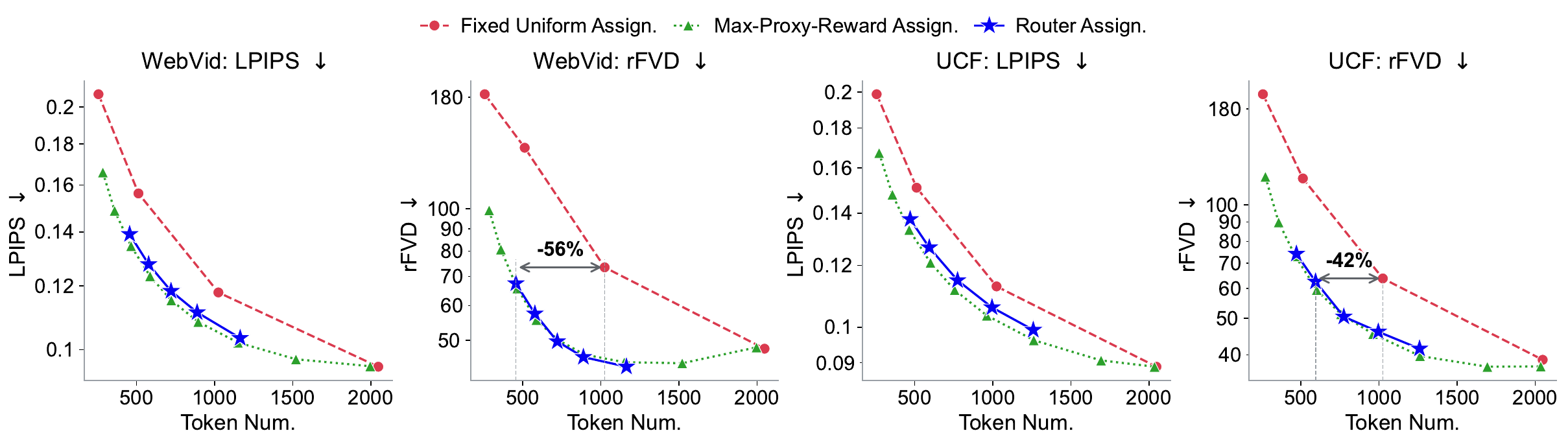}
    % \vspace{-0.10in}

    \caption{\textbf{Quality-cost trade-off curves for different assignment strategies.} By adaptively assigning token budgets to different temporal blocks across various videos, our max-proxy-reward strategy~(green series) achieves superior performance under various overall budgets compared to the typical fixed uniform token assignment approach~(red series). The router-based assignment~(blue series) delivers performance close to that of the max-proxy-reward strategy on both WebVid and UCF datasets~(the latter unseen during router training).}
    
    % \caption{\textbf{Quality-cost trade-off curves for different assignment strategies.} By adaptively assigning token budgets to different temporal blocks of different videos, our max-proxy-reward strategy~(green series) achieves better performances under different overall budgets, compared to the typical approach of fixed uniform token assignment~(red series). Applying routers for assignments~(blue series) achieves good performance close to the max-proxy-reward assignment, on both WebVid and UCF~(which is not seen in router training).  }
    \label{fig:trade_off_curve}
    \vspace{-0.10in}

\end{figure*}

\subsection{Settings}
% \tianwei{Need to put the very specific details in supp.}
\noindent\textbf{Dataset.} We apply the commonly used combination of UCF~\cite{Soomro2012ucf101} and K600~\cite{Carreira2018K600} datasets for video reconstruction and generation experiments. And for validation on more general data, we additionally experiment on WebVid-10M~\cite{Bain21WebVid} for video reconstruction. In all experiments, we use $16\times128\times128$ video clips for training and evaluation. For router training, the video data is a randomly sampled subset of WebVid-10M, containing 100k video clips. 

\noindent\textbf{Implementation details.} When patchifying videos in \ours{}, the spatial downsample ratio is 8 and the temporal downsample ratio is 4. Therefore, a $16\times128\times128$ video produces $4\times16\times16$ features. When initializing 1D query embeddings, the candidate length of each temporal block is in $\{512, 256, 128, 64, 32\}$, so the number of all candidate assignments is $5^4=625$. \ours{} applies a codebook size of 16384 by default. But for the final tokenizers trained on UCF and K600 dataset, we use 8192 codebook size for fair comparison to previous methods~\cite{larp, Li2025AdapTok}. We train 19.9M ViT-S size routers in stage 3. 
% In its training data curation, for proxy reward calculation in Eq.~\ref{eq:proxy reward}, we calculate $Q(\mathcal{E}_\text{proxy}, x, a)$ as the normalized LPIPS~\cite{LPIPS} between $x$ and the reconstructed video, and $L(a)$ as the normalized length. 
We train Llama-like~\cite{llama, llamagen} GPT models on variable length sequences. For class2video generation, the condition token corresponds to class labels, and for K600 frame prediction the conditions are tokens encoded from 8 frames padded from the 5 condition frames. Per-frame reconstruction fidelity metric LPIPS~\cite{LPIPS} and the overall distribution fidelity metric FVD~\cite{unterthiner2018fvd} are used for evaluation.
More details on the training and inference of the AR models can be found in the supplementary material.

\subsection{Validation on Quality-Cost Trade-off Curves}

We validate the effect of our max-proxy-reward searching strategy and routers on proxy tokenizers. We compare these approaches against a commonly used baseline: fixed uniform assignment, which allocates the same number of tokens to different temporal blocks across videos. We plot quality-cost trade-off curves to show the overall trends of these assignment strategies under varying overall budgets.

% In this part, we aim to validate the effect of our max-proxy-reward searching strategy and routers on proxy tokenizers. We compare the two approaches to a commonly used baseline: uniform assignment which allocates the same number of tokens to different temporal blocks of different videos. We plot the quality-cost trade-off curve to show the overall trend of different strategies under different overall budgets.

To generate the quality-cost curves, we adjust the overall budgets to obtain multiple evaluation points. For the uniform assignment strategy, we vary the number of tokens allocated to each temporal block from 64 to 512. For the max-proxy-reward assignment strategy, we adjust $w_q$ in steps of 0.2 from 0.4 to 2.0, while setting $w_l = 2.0 - w_q$, thereby producing various $w_q, w_l$ combinations that alter the overall budgets. For the router assignment strategy, we employ routers trained under different $w_q, w_l$ combinations: $w_q$ ranging from 0.8 to 1.6 in steps of 0.2, with $w_l = 2.0 - w_q$. We evaluate on two benchmarks: the WebVid validation set and the UCF-101 training set, using two corresponding proxy tokenizers trained on either the WebVid training set or the UCF \& K600 training sets.

As shown in Fig.~\ref{fig:trade_off_curve}, on both datasets, the max-proxy-reward assignment strategy yields a quality-cost curve that achieves superior LPIPS and reconstruction FVD~(rFVD) at equivalent overall budgets. Moreover, the routers closely align with the quality-cost curve of the max-proxy-reward strategy, demonstrating their ability to effectively simulate it. The routers also generalize well to different proxy tokenizers and datasets not seen during training, as evidenced in the last two columns of Fig.~\ref{fig:trade_off_curve}. The routers significantly reduce the total budget while delivering even better overall reconstruction quality. Focusing solely on rFVD, compared to traditional fixed uniform assignment approaches that allocate 1024 tokens on average to a $16\times 128\times128$ video clip, the routers can achieve 56\% token savings on WebVid and 42\% on UCF, with comparable or even better performance. We validate the performance and generalization of the routers on proxy tokenizers. Next, we demonstrate the advantages of using routers to guide final tokenizer training.

% As shown in Fig.~\ref{fig:trade_off_curve}, on both datasets, the max-proxy-reward assignment strategy presents a quality-cost curve that can achieve better LPIPS and rFVD under the same overall budgets. And the routers can closely fit into the quality-cost curve of the max-proxy-reward assignment strategy, showing that they can well simulate the max-proxy-reward strategy. Also, the performances of the routers can generalize to a different proxy tokenizer on a different dataset, which are not involved in their training, as shown in the last two columns in Fig.~\ref{fig:trade_off_curve}. The routers can help significantly decrease the overall budget while achieving potentially even better reconstruction quality. If we focus only on rFVD, compared to typical approaches which allocate 1024 tokens on average to a $16\times 128\times128$ video clip, using routers can save $56\%$ of tokens on WebVid and $42\%$ on UCF.

% So far, we have validated the performance and generalization of the routers on proxy tokenizers. Next, we demonstrate the advantages of using routers to guide final tokenizer training.

% So far, on proxy tokenizers, we have validated the performance and generalization of the routers. Next we will further show the advantage of using routers to guide the final tokenizer training.  

\begin{table}[]
    \centering
    \resizebox{0.5\textwidth}{!}{
        \begin{tabular}{lcccc}
        \toprule
         Settings
         & \textbf{PSNR$\uparrow$} & \textbf{LPIPS$\downarrow$} & \textbf{rFVD$\downarrow$} &  \textbf{\#rTokens$\downarrow$} \\
        \midrule
         A1. Uniform~(Proxy Tok.)      & 27.26 & 0.1178    & 73    &   1024  \\
        
           A2.     Uniform~(Final Tok.)      & 27.77 & 0.1056     & 63   &   1024 \\
                 A2 + VideoMAE Disc.            & 26.68 &   0.1197  &   13  &  1024   \\
        \midrule
         B1. Router~(Proxy Tok.)            & 27.05 & 0.1182    & 50    &  721\small{(-29.6\%)} \\

        B2. Router~(Final Tok.)            & 27.68 & 0.1068    & 33    &   721\small{(-29.6\%)}  \\
         B2 + VideoMAE Disc.            & 26.90 &   0.1144  &  9.2   &  721\small{(-29.6\%)}   \\
        \bottomrule
        \end{tabular}
    }
    \caption{\textbf{Final tokenizer validation on WebVid.} The tokenizers are trained for 400k iterations. With the router, final tokenizers achieve comparable LPIPS and better rFVD with 29.6\% saving in token length~(row A2 \vs B2 and row A2+ \vs B2+). Final tokenizers outperform proxy tokenizers with the same training efforts~(row A2 \vs A1, B2 \vs row B1), showing the importance of bridging the training-inference gap for variable-length tokenizers.
    \label{tab:final_validation}
    }    
    \vspace{-0.10in}
\end{table}
\begin{table}[]
    \centering
    \resizebox{0.5\textwidth}{!}{
        \begin{tabular}{lccccc}
        \toprule
         Settings & \textbf{LPIPS$\downarrow$} & \textbf{rFVD$\downarrow$} & \textbf{\#rTokens$\downarrow$}&  \textbf{gFVD$\downarrow$} &
         \textbf{\#gTokens$\downarrow$}
         \\
        \midrule
        Uniform~(Final)      & 0.1303 & 13     & 1024   &   98 & 1024\\
         Router~(Final)  & 0.1212 & 13    & 774\small{(-24.4\%)}   &  96 & 740\small{(-27.7\%)} \\
        \bottomrule
        \end{tabular}
    }
    \caption{\textbf{Final tokenizer validation on UCF.} The final tokenizer with router beats the fixed uniform assignment baseline in both reconstruction and downstream AR generation, while saving 24.4\% and 27.7\% token length separately.
    \label{tab:final_validation_ucf}
    }
    \vspace{-0.10in}

\end{table}
\begin{table*}[t!]
    \centering
    % \resizebox{\textwidth}{!}
    {
        \begin{tabular}{lccccccc}
        \toprule
        \textbf{Method} & \multicolumn{2}{c}{\textbf{\#Params}} & \textbf{\#rTokens} & \textbf{rFVD}$\downarrow$ & \multicolumn{2}{c}{\textbf{gFVD}$\downarrow$} & \textbf{\#gTokens} \\
         \cmidrule{2-3} \cmidrule{6-8} 
         & \small{Tokenizer} & \small{Generator} &  &  & K600  & UCF &  UCF \\
        \midrule
        \multicolumn{7}{l}{\textcolor{gray}{\textit{Diffusion-based generative models with continuous video tokenizers}}} \\
        \hdashline
        \addlinespace
        \textcolor{gray}{VideoFusion}~\cite{luo2023videofusion} & \textcolor{gray}{-} & \textcolor{gray}{2B} & \textcolor{gray}{-} & \textcolor{gray}{-} & \textcolor{gray}{-} & \textcolor{gray}{173} & \textcolor{gray}{-}  \\
        \textcolor{gray}{HPDM}~\cite{skorokhodov2024hdpm} & \textcolor{gray}{-} & \textcolor{gray}{725M} & \textcolor{gray}{-} & \textcolor{gray}{-} & \textcolor{gray}{-} & \textcolor{gray}{66} & \textcolor{gray}{-}  \\
        \textcolor{gray}{W.A.L.T-L}~\cite{gupta2024walt} & \textcolor{gray}{-} & \textcolor{gray}{313M} & \textcolor{gray}{-} & \textcolor{gray}{-} & \textcolor{gray}{3.3} & \textcolor{gray}{46} & \textcolor{gray}{-}  \\
        \midrule
        % VideoFusion~\cite{luo2023videofusion} & - & 2B & - & - & - & 173 & -  \\
        % HPDM~\citep{skorokhodov2024hdpm} & - & 725M & - & - & - & 66 & -  \\
        % W.A.L.T-L~\cite{gupta2024walt} & - & 313M & - & - & 3.3 & 46 & -  \\
        % \midrule
        \multicolumn{7}{l}{\textit{MLM generative models with discrete video tokenizers}} \\
        \hdashline 
        \addlinespace
        MAGVIT-MLM~\citep{yu2023magvit} & 158M & 306M & 1024 & 25 & 9.9  &  76 & 1024 \\
        MAGVIT-v2-MLM~\citep{magvit-v2} & - & 307M & 1280 & \textbf{8.6} & \underline{4.3} & 58 & 1280 \\
        \midrule
        \multicolumn{7}{l}{\textit{AR generative models with discrete video tokenizers}} \\
        \hdashline 
        \addlinespace
        CogVideo~\citep{hong2022cogvideo} & - & 9.4B & - & - & 109.2 & 626 & - \\
        TATS~\citep{ge2022TATS}           & 32M & 321M & 1024 & 162 & - & 332  & 1024 \\
        MAGVIT-AR~\citep{yu2023magvit}    & 158M & 306M & 1024 & 25 & - & 265 & 1024 \\
        MAGVIT-v2-AR~\citep{magvit-v2} & - & 840M & 1280 & \textbf{8.6} & - & 109 & 1280 \\
        OmniTokenizer~\citep{wang2024omnitokenizer} & 82.2M & 650M & 1280 & 42 & 32.9 & 191 & 1280 \\
        AdapTok~\cite{Li2025AdapTok} & 195M & 633M & 1024 & 36 & 11  & 67 & 1024 \\
        % \rowcolor{gray!15}
        % \rowcolor{gray!15}
        LARP-L-Long~\cite{larp} & 173M & 343M & 1024 & 20 & 6.2  &102 & 1024 \\
        % \rowcolor{gray!15}
        LARP-L-Long~\cite{larp} & 173M & 632M & 1024 & 20 & 5.1  & \underline{57} & 1024 \\
        \rowcolor{gray!15}
       \ours{}~(ours) & 145M & 327M & 774\small{(-24.4\%)} & \underline{9.7} & 4.6  & 62 & 756\small{(-26.2\%)} \\
        \rowcolor{gray!15}

        \ours{}~(ours) & 145M & 633M & 774\small{(-24.4\%)} & \underline{9.7} & \textbf{4.0}  & \textbf{48} & 756\small{(-26.2\%)} \\

        \bottomrule
        \end{tabular}
    }
    \caption{
    \textbf{System-level comparison for tokenizers and downstream generation models.} \ours{} achieves superior performances in UCF-101 video reconstruction, downstream class-to-video generation and K600 frame prediction, while saving 24.4\% tokens in reconstruction and 26.2\% tokens in UCF class-to-video generation.
    % Results are grouped by the type of generative models.
    % The scores for MAGVIT-AR and MAGVIT-v2-AR are taken from the appendix of MAGVIT-v2~\citep{yu2023language}. \systemm-L-Long denotes the \systemm-L trained for more epochs. 
    % Our best results are obtained with a larger AR generator.
    \label{tab:comp}
    }
    % \vspace{-0.10in}

\end{table*}

\subsection{Validation on Final Adaptive Tokenizer }
\label{sec: final validation}

% goal (motivation)
We validate the benefits of using routers to provide optimal assignment guidance in both the training and inference of adaptive video tokenizers, which bridges the training-inference gap in previous adaptive length video tokenizers~\cite{yan2025elastictok, Li2025AdapTok}. For the choice of router, we choose the router conditioned on $w_q=1.2, w_l=0.8$, which achieves comparable LPIPS and better rFVD with $24.4\%$ token length saving against the uniform assignment. We conduct two sets of experiments on WebVid-10M and the combination of UCF \& K600 separately. 

On the WebVid-10M dataset, with the assignments provided the a router, we train two variants of final tokenizers, one not using the VideoMAE discriminator while the other uses it. As baselines, with fixed uniform assignments of 1024 tokens per video, we train two variants of final tokenizers correspondingly. And all tokenizers evaluated in this experiment are trained for 400k iterations for fair comparison. As shown in Tab.~\ref{tab:final_validation}, on the WebVid validation set, we validate that for final tokenizers, the router-guided tokenizer achieves comparable LPIPS, better rFVD and $29.6\%$ token length savings against the tokenizer trained with uniform assignments, no matter using VideoMAE discriminator or not~(row A2 \vs B2 and A2+ \vs B2+). Besides, Tab.~\ref{tab:final_validation} also indicates that the final tokenizers outperform the proxy tokenizers with the same training iterations~(row A2 \vs A1 and B2 \vs B1), proving that using optimal assignments in both tokenizer training and inference benefits performances.

On UCF \& K600 dataset, we train two tokenizers with fixed uniform assignment or router-guided assignment, both using the VideoMAE discriminator. As shown in Tab.~\ref{tab:final_validation_ucf}, under the predicted optimal assignment, the final tokenizer trained with the router achieves even better reconstruction with 24.4\% savings in token length. Further, we train 99M GPT-B AR generation models on the two tokenizers separately on the UCF-101 class2video dataset, and evaluate them by generation FVD~(gFVD) based on 10k generated samples. The AR model that adaptively decides the length of tokens it generates achieves even better gFVD while generating 740 tokens per video on average, saving 27.7\% of tokens compared to the fixed-length AR model.

By training and evaluating final tokenizers with the adaptive assignments of our router, we show that a router helps train a better adaptive tokenizer by eliminating the previous training-inference gap, beating baselines trained with fixed uniform assignment. And importantly, for the first time, we show that downstream AR models trained on adaptive length video token sequences can achieve better overall generation quality with significant savings in token length cost.

\begin{table}[b]
    \centering
    % \vspace{-0.15in}
    \resizebox{0.5\textwidth}{!}{
        \begin{tabular}{lcccc}
        \toprule
        \textbf{Methods} & \textbf{Tok. Param.} & \textbf{\#gTokens} & \textbf{AR Param.} & \textbf{gFVD$\downarrow$} \\
        \midrule
        AdapTok~\cite{Li2025AdapTok} & 195M & 1024 & 633M  &  11\\
        LARP~\cite{larp} & 173M & 1024 & 632M  &  5.1\\
        \rowcolor{gray!15}
        \ours{}~(ours) & 145M & 862~\small{(-15.8\%)} & 633M & \textbf{4.0} \\
        \bottomrule
        \end{tabular}
    }
    \caption{\textbf{K600 frame prediction comparison.} In similar settings, \ours{} performs the best with 15.8\% less generated tokens.
    }
    \label{tab:k600_comp}
    % \vspace{-0.10in}
\end{table}

\subsection{System-Level Comparison}

We compare \ours{} with previous video generative models, evaluating performance in terms of video reconstruction, generation quality, and average token length. These aspects are assessed using the UCF-101 reconstruction, UCF-101 class-to-video generation, and K600 frame prediction benchmarks. As shown in Tab.~\ref{tab:comp}, \ours{} achieves substantially better reconstruction FVD~(rFVD) while reducing token length by 24.4\% compared to LARP~\cite{larp}. \ours{} establishes a new state-of-the-art~(SOTA) on UCF-101 class-to-video generation, with a generation FVD~(gFVD) of 48 and 26.2\% fewer tokens than the previous SOTA method, LARP~\cite{larp}. \ours{} also delivers the best results on K600 frame prediction. 
For a fair comparison of generation efficiency on K600 frame prediction, we benchmark against AdapTok~\cite{Li2025AdapTok} and LARP~\cite{larp} in Tab.~\ref{tab:k600_comp}, as we employ the same approach of additionally generating conditioning frames during both training and inference. We fix the token length for conditioning frames as $512+128=640$, therefore, during frame prediction training, we choose assignments with $(512,128)$ prefix for the first two temporal blocks that have the highest probability predicted by the router, for the encoding of 16-frame samples. Under this setting, \ours{} achieves the best gFVD on K600 frame prediction while saving 15.8\% in generated tokens.

We demonstrate that adaptive video tokenization, combined with our advanced training recipe, enables \ours{} to achieve both high efficiency and leading performance on video reconstruction and downstream AR generation.

\subsection{Ablation Study}
\label{sec:ablation}

\begin{figure}[t]
% \vspace{-0.1in}
    \centering
    \includegraphics[width=\linewidth]{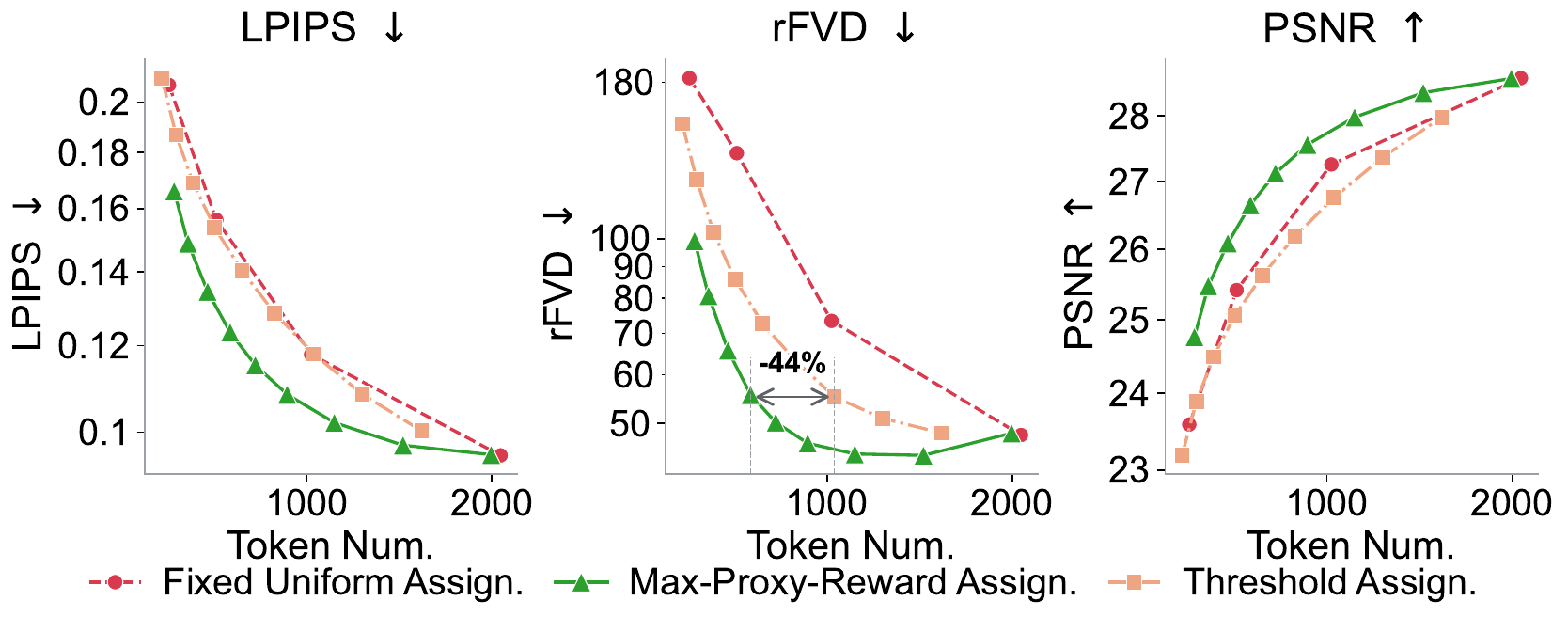}
    \caption{\textbf{Quality-cost curve: threshold based \vs max-proxy-reward \vs uniform assignment.} While threshold-based assignment improves rFVD against uniform assignment, it underperforms our max-proxy-reward strategy.}
    \label{fig:trade_off_curve_ablation}
    % \vspace{-0.10in}

\end{figure}

\noindent\textbf{Threshold \vs max-proxy-reward searching.} For adaptive video tokenization, ElasticTok~\cite{yan2025elastictok} applies a heuristic method to find adaptive assignments: searching the minimum token length to be kept that maintains the reconstruction quality above certain thresholds. This heuristic threshold-based method is not designed to optimize for the overall quality-cost trade-off. To compare this threshold-based method to our max-proxy-reward strategy, we implement a similar baseline method in our setting, which finds the assignment with the minimum token length and satisfies a certain LPIPS threshold for each video. If no assignment can satisfy the threshold, the maximum length assignment will be chosen. Testing on our proxy tokenizer, by varying the LPIPS threshold, we can plot the quality-cost trade-off curve for this baseline strategy. As shown in Fig.~\ref{fig:trade_off_curve_ablation}, on the WebVid validation set, while the threshold-based baseline improves the rFVD compared to uniform assignment, it still lags behind our max-proxy-reward strategy.

% ElasticTok~\cite{yan2025elastictok} chooses the assignments for input videos by searching for the minimal length assignments among all assignments that satisfy the reconstruction quality threshold. 

% We implement a similar searching strategy as the baseline to compare with our max-proxy-reward assignment searching strategy on our proxy tokenizer. Specifically, in this baseline strategy, for each video, the assignment allocated to it will have the shortest token length among assignments that lead to an LPIPS value lower than a specified threshold. By varying the LPIPS threshold, we can plot the quality-cost trade-off curve for this baseline strategy. As shown in Fig.~\ref{fig:trade_off_curve_ablation}, on WebVid validation set, while the threshold-based baseline improves the rFVD compared to uniform assignment, it still lags behind our max-proxy-reward strategy.  

% Shown in Fig.~\ref{fig:trade_off_curve_ablation}. We compare our max-proxy-reward searching to the threshold-based searching from ElasticTok~\cite {yan2025elastictok} in terms of overall performance in the quality-cost trade-off curve.

\begin{table}[]
    \centering
% \vspace{-0.10in}
    \resizebox{0.5\textwidth}{!}{
        \begin{tabular}{lccc|c}
        \toprule
        \textbf{Configuration} & \textbf{PSNR$\uparrow$} & \textbf{LPIPS$\downarrow$} & \textbf{rFVD$\downarrow$} & \textbf{gFVD$\downarrow$} \\
        \midrule
        Final Recipe~(Uniform)             & 25.05 & 0.1303    &   \textbf{13}  & \textbf{98} \\
        \midrule
        - VideoMAE~\cite{tong2022videomae, wang2022internvideo} Disc.      & 26.21 & 0.1097     & 65    &   155 \\
        - V-JEPA2~\cite{vjepa2} Align. & 25.30 & 0.1253    & 18    &   144 \\
        - Both & 26.41 &  0.1095  &  80   &  230 \\
        % + Unifrom$\rightarrow$Router & 25.42 &    0.1212  &  13   &  96 \\

        \bottomrule
        \end{tabular}
    }
    \caption{\textbf{Ablation study for video representation alignment and video semantic discriminator.} Removing either design will lead to degradation in rFVD and downstream gFVD.
    }
    \label{tab:tokenizer_ablation}
        % \vspace{-0.10in}

\end{table}

% Video semantic encoders are like strong teachers; they can help video tokenizers like a teacher guiding a student: they assist through representation alignment, much like demonstrating problem-solving, and through acting as discriminators, akin to grading and providing feedback on the student's answers.

\noindent\textbf{Video semantic encoder for video tokenizers.} 
Video semantic encoders can help video tokenizer training in two ways:~(1) providing representation alignment;~(2) giving perceptual feedback as the discriminator for better reconstruction quality. We study the two designs and reveal that they are both important for reconstruction and downstream generation for video tokenizers. Here, for simplicity, we utilize the typical uniform assignment and train video tokenizers under different recipes on the UCF\&K600 dataset for 400k iterations, and evaluate them by reconstruction and downstream GPT-B model generation on UCF-101. As shown in Tab.~\ref{tab:tokenizer_ablation}, removing either representation alignment or VideoMAE discriminator can lead to degradation of rFVD and downstream gFVD. While we do recognize a drop in the per-frame fidelity metric PSNR and LPIPS brought especially by the VideoMAE discriminator, in our qualitative checking, we identify that the drop in PSNR and LPIPS is traded for less blurriness and less temporal flickering, which corresponds to higher rFVD. More qualitative comparisons are in the supplementary material.

\section{Conclusions}
\label{sec:conclusions}

% In this work, we propose \ours{}, a content-adaptive tokenization framework that efficiently allocates tokens across different temporal blocks and samples. We identify a key challenge: the absence of a utility function for identifying optimal assignments that achieve the best quality-cost trade-off. To address this, we introduce the concept of proxy rewards and reformulate efficient assignment selection as a maximum proxy reward assignment classification task. This formulation enables us to train routers that directly predict optimal assignments for input videos based on specified quality-cost preference weights. Consequently, these routers can be used to dynamically allocate token lengths across samples and time blocks during the training of tokenizers and downstream autoregressive (AR) generation models. Further augmented by video semantic encoders as feature alignment teachers and perceptual discriminators, \ours{} not only establishes new state-of-the-art performance in UCF-101 class-conditioned video generation and Kinetics-600 frame prediction but also reduces generated token lengths by 26.2\% and 15.8\%, respectively. For discussions on limitations and future work, please refer to the supplementary materials.

In this work, we propose \ours{}, a content-adaptive video tokenization framework to efficiently assign tokens across different temporal blocks and videos. We introduce proxy reward as a novel metric for finding the optimal assignments with the best quality-cost trade-off. By reformulating optimal assignment selection for video tokenization as a maximum proxy reward assignment classification task, we can curate supervision datasets to train routers to map each video to its optimal assignment. These routers help us train video adaptive tokenizers and downstream autoregressive~(AR) video generative models with efficient token assignments. Enhanced by our advanced recipe incorporating video semantic encoders in tokenizer training, \ours{} achieves superior reconstruction and downstream AR generation quality while significantly saving token length cost. \ours{} has presented promising results on 16-frame videos in this work, and for future development, can potentially achieve higher efficiency for videos with longer duration. Please refer to the supplementary material for more discussions on future work and limitations of \ours{}.

% videos with longer duration potentially redundancy 

% We identify a core problem of missing utility for finding the optimal assignments with the best quality-cost trade-off. By introducing the concept of proxy reward, we formulate efficient assignment selection as a max-proxy-reward assignment classification task. This task formulation allows us to train routers that directly predict optimal assignments for input videos under specified quality-cost preference weights. Therefore, we can use the routers to efficiently allocate token lengths for different samples across different time blocks during the training of tokenizers and downstream AR generation models. Further enhanced with video semantic encoders as the feature alignment teacher and perceptual discriminator, \ours{} not only sets new state-of-the-art generation performances in downstream UCF-101 class2video AR generation and K600 frame prediction but also achieves 26.2\% and 15.8\% saving in generated token length accordingly. 

\clearpage

\section*{Acknowledgments}

This work is supported in part by the Research Grant Council of Hong Kong through the NSFC-RGC Joint Research Scheme under grant N\_HKU769/25.

{
    \small
    \bibliographystyle{ieeenat_fullname}
    \bibliography{main}
}

\clearpage
\setcounter{page}{1}
\maketitlesupplementary
\renewcommand{\thesection}{\Alph{section}}

% \tableofcontents
\section*{Content of the Appendix}
This supplementary material includes the following content:
\begin{itemize}
  \item Sec.~\ref{sec:limitations} discusses the limitations of \ours{}.
   \item Sec.~\ref{sec:future work} provides the plan for the future work.
  \item Sec.~\ref{sec:implementation} gives the detailed implementation of the four-stage framework of \ours{} and downstream adaptive length AR models.
  \item Sec.~\ref{sec:qual_analysis} provides the reconstruction performances of our final tokenizers, including qualitative examples for the adaptive length reconstruction and generation, as well as cases of how VideoMAE discriminator affects video reconstruction perceptually.
  \item Sec.~\ref{sec:compute_analysis} provides the compute cost analysis for our four-stage framework and downstream AR generation model training.
  \item Sec.~\ref{sec:router_analysis} analyzes the router's max-proxy-reward assignment predictions in terms of accuracy and proxy reward.
  \item Sec.~\ref{sec:attn mask} explains the attention mask mechanism in our Q-Former style video adaptive tokenizers.
  \item Sec.~\ref{sec:img adaptive exp} shows the results for translating the solution of \ours{} to image adaptive tokenization.
  
\end{itemize}

\section{Limitations}
\label{sec:limitations}

In this work, we focus on addressing the key challenge in adaptive length video tokenization: identifying the optimal assignment. Although we have demonstrated the superiority of our method in video reconstruction, as well as downstream autoregressive (AR) class-to-video generation and frame prediction tasks, our experiments were limited to $16 \times 128 \times 128$ video clips. We did not evaluate it on videos with higher resolution or longer duration that align with industry-level requirements~\cite{wan2025, cui2025emu35}. Additionally, due to limited computational resources, we have not validated \ours{} on more complex downstream tasks, such as text-to-video generation.

Furthermore, when extending video duration in adaptive length video tokenizers, the number of possible assignment choices can grow exponentially if exhaustive searching is naively applied. Although this issue is not addressed in the current work, we discuss a potential solution in our future work section (Sec.~\ref{sec:future work}), which can reduce the complexity of optimal assignment searching from $O(m^t)$ to around $O(t^2)$ with respect to the maximum video duration $t$.

% In this work, we focus on resolving the key issue for adaptive length video tokenization, which is optimal assignment identification. Although we have validated the superiority of our method on video reconstruction and downstream AR class-to-video generation and frame prediction tasks, we only experiment on $16\times 128\times128$ video clips, instead of on videos with higher resolution and longer duration aligned to industrial level needs~\cite{wan2025, cui2025emu35}. We also have not validated \ours{} on more complex downstream tasks like text-to-video generation, due to limited computational resources.  Besides, when extending the video duration for adaptive length video tokenizers, the number of possible choices of assignments can increase exponentially if we naively conduct exhaustive searching for all possible choices. Although this issue is not addressed in the work, in our future work discussion~(Sec.~\ref{sec:future work}), we present a potential solution that can simplify the complexity of optimal assignment searching into linear complexity with respect to maximum video duration.

% we only explore adaptive video tokenization on $16\times 128\times128$ clips due to computational resource constraints, while in real-world applications, the duration and spatial resolution of videos will be larger. 

\section{Future Work}
\label{sec:future work}

\noindent\textbf{Adaptive length video tokenization on longer videos.} In the main paper, when identifying the optimal assignment for a video clip with $T$ temporal blocks and $m$ possible token number choices for each temporal block, we search for $m^T$ possible assignments to find the optimal one. This approach will become unaffordable for larger $T$. To address this, in future work, we will explore a method that searches for optimal assignments approximately in an autoregressive way. For example, for a video with $2T$ temporal blocks, we can first search the $m^T$ possible assignments for the first $T$ blocks, then, based on the optimal assignment for the first $T$ blocks, we continue to search the $m^T$ possible assignments for the $T$ blocks. Therefore, if we assume the reconstruction cost for the proxy tokenizer increases linearly with longer $T$, then the complexity for optimal assignment searching is estimated to be $O(T^2)$. 

\noindent\textbf{Extension to adaptive length video VAE and diffusion models.} The idea of adaptive length video tokenization is not limited to discrete tokenizers, and can naturally transfer to adaptive length VAE~\cite{vae, vae_journal} training. While it is natural for AR models to learn on variable length sequences, the performance of diffusion models on denoising adaptive length sequences can be discussed in future work.

\noindent\textbf{Router improvements.} In our current implementation, the preference weights are implicitly fixed during the process of training data curation for routers. In the future work, we may want the preference weights to be able to be input explicitly for the routers for more flexible applications.

\section{Implementation Details}
\label{sec:implementation}

\noindent\textbf{Tokenizer training.} On WebVid-10M, we train variable length tokenizer on 3 FPS video frames, following the approaches in VidTok~\cite{tang2024vidtok}. On UCF \& K600 dataset, we train tokenizers on video frames with their original FPS, following typical settings. We use a cosine learning rate schedule. The maximum learning rate is $1\times 10^{-4}$ and end learning rate is $1\times 10^{-6}$. The batch size is 128, and proxy tokenizers are all trained for only 400k iterations before being used for proxy reward calculation. The final video adaptive tokenizers are trained for 1000k iterations, whose training cost is aligned with previous work~\cite{larp, Li2025AdapTok}.

\noindent\textbf{Proxy reward calculation.} For proxy reward calculation from the main paper:
\begin{equation}
\label{eq:supp_proxy reward}
    R_\text{proxy} = w_q Q(\mathcal{E}_\text{proxy}, x, a) - w_l L(a) 
\end{equation}
Specifically, we calculate $Q(\mathcal{E}_\text{proxy}, x, a)$ as:
\begin{equation}
    Q(\mathcal{E}_\text{proxy}, x, a) =
    \frac{\text{LPIPS}(\mathcal{E}_\text{proxy}(x, a), x) - \text{MEAN}_{\text{LPIPS}}}{\text{STD}_{\text{LPIPS}}}
\end{equation}
where $\text{LPIPS}(\mathcal{E}_\text{proxy}(x, a), x)$ is the LPIPS value between original video $x$ and the reconstruction result $\mathcal{E}_\text{proxy}(x, a)$ using proxy tokenizer $\mathcal{E}_\text{proxy}$ under assignment $a$. $\text{MEAN}_\text{LPIPS}$ denotes the expectation of $\mathcal{E}_\text{proxy}(x, a)$ for randomly sampled $x$ from all the training videos and randomly sampled $a$ from all candidate assignments, and $\text{STD}_{\text{LPIPS}}$ represents the standard deviation of $\mathcal{E}_\text{proxy}(x, a)$ for random $x$ and $a$. We choose LPIPS for per-video reconstruction quality measurement, as it is a metric designed to better align with human perception. We calculate $L(a)$ as:
\begin{equation}
    L(a) = \frac{\sum_{k=1}^T a[k] - \text{MEAN}_L}{\text{STD}_{L}}
\end{equation}
where $\sum_{k=1}^T a[k]$ is the sum of the allocated tokens across all $T$ temporal blocks. $\text{MEAN}_L$ and $\text{STD}_{L}$ are the expectation of $\sum_{k=1}^T a[k]$ for randomly sampled $a$.

\noindent\textbf{Router training.} We train 19.9M ViT-S size routers with a batch size of 128 for 50k iterations. We optionally use frozen V-JEPA2~\cite{vjepa2} to patchify raw frames into video embeddings. Otherwise, we use the typical learnable linear projection for patch embeddings. In practice, we find that there is no obvious performance gap between these two visual embedding strategies.

\noindent\textbf{AR model training.} For the adaptive length token sequences produced by \ours{}, before each temporal block, a special token indicating the number of tokens for the upcoming temporal block will be inserted for AR training. Therefore, for AR inference, before generating the tokens of the next temporal block, the AR model first predicts the special tokens indicating the length of the next block. On UCF-101, AR models are trained for 3000 epochs using WSD~\cite{wsd_scale} learning rate schedule, where the learning rate is kept constant for the first 80\% of the training and quickly annealed to 0 in the rest 20\% training iterations. On K600, the AR models are trained for 75 epochs with the same learning rate scheduler.

\noindent\textbf{AR inference for adaptive length video generation.} In adaptive length AR generation, we observe that even with a special token preceding each temporal block to indicate the number of tokens in the upcoming block, the AR model may still occasionally produce unexpected tokens during inference~(\eg, sampling a special token when a visual token is expected, or vice versa). To ensure the model generates precisely the number of tokens specified by the preceding special token for each temporal block, we employ a logit-masking strategy. For instance, when sampling the first token of a variable length sequence, which is expected to be a special token denoting the token count for the initial block, all logit entries corresponding to visual tokens are masked to \(\tt -inf \) before the softmax operation. This guarantees that only a special token is sampled. Subsequently, for the next 
$k$ tokens~(as indicated by the special token), the logits for special tokens are masked, ensuring only visual tokens are generated. This process continues until $m$ special tokens and their corresponding temporal blocks are generated. This approach incurs nearly no additional computational overhead and guarantees the generated variable length sequence maintains the correct structure. We use constant classifier free guidance~(CFG) schedules for AR model inference during class-to-video sampling. For GPT-B models, the CFG value is $2.5$, for larger GPT models, we use $1.75$ CFG value.

\section{More Results and Qualitative Analysis}
\label{sec:qual_analysis}
\begin{table}[]
    \centering
    \resizebox{0.5\textwidth}{!}{
        \begin{tabular}{lcccc}
        \toprule
         Settings / Dataset
         & \textbf{PSNR$\uparrow$} & \textbf{LPIPS$\downarrow$} & \textbf{rFVD$\downarrow$} &  \textbf{\#rTokens$\downarrow$} \\
        \midrule        
        Final w/ VideoMAE Disc.~(WebVid)            & 27.37 &   0.1063  &   7.3  &  721  \\
        Final w/o VideoMAE Disc.~(WebVid)            & 28.18 &   0.0983  &   32  &  721   \\
        \midrule
        Final w/ VideoMAE Disc.~(UCF\&K600)            & 25.75 &   0.1140  &   9.7  &  774   \\ 
        \bottomrule
        \end{tabular}
    }
    \caption{\textbf{The detailed performances of final tokenizers reconstruction results.} All models are trained for the full 1000k iterations. The tokenizers trained on WebVid-10M are evaluated on the WebVid validation set, and the tokenizers trained on UCF-101 and K600 are evaluated on UCF-101 training set.
    }    
    \label{tab:reconstruction_final}
    \vspace{-0.10in}
\end{table}

\begin{figure*}[]
% \vspace{-0.1in}
    \centering
    \includegraphics[width=\linewidth]{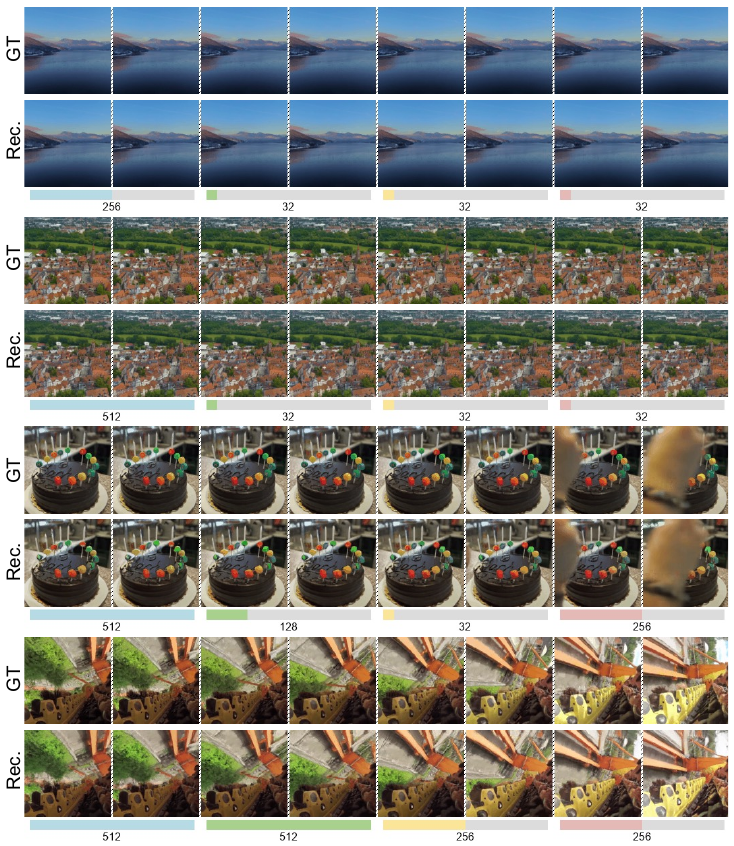}
    % \vspace{-0.30in}
    \caption{\textbf{Adaptive reconstruction results on WebVid.} We downsample 16 frames into 8 frames for visualization, and each two frames represent a 4-frame temporal block.
    The router typically assigns more tokens to the initial temporal block, allowing the reconstruction of subsequent frames to also benefit from more precise information encoded for the initial block. Content with simple layouts receives fewer tokens~(first example \vs other examples). If later frames largely repeat previous ones, they are assigned the minimum number of tokens. Video clips that vary constantly and intensely are allocated more tokens.
    }
    \label{fig:qual_grid_webvid}
\end{figure*}
\begin{figure*}[t]
% \vspace{-0.1in}
    \centering
    \includegraphics[width=\linewidth]{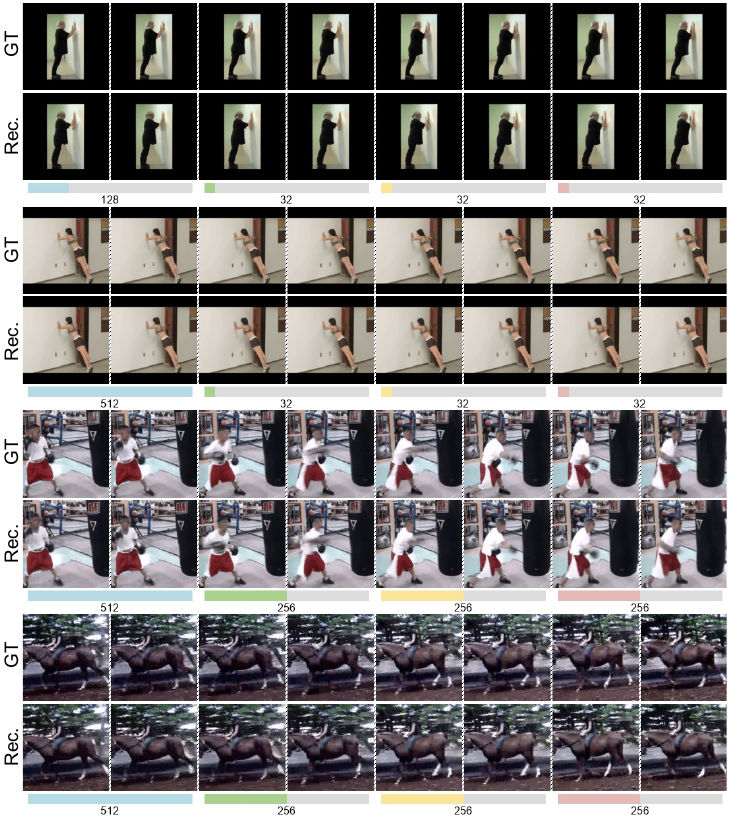}
    % \vspace{-0.30in}
    \caption{\textbf{Adaptive reconstruction results on UCF-101.} We downsample 16 frames into 8 frames for visualization, and each two frames represent a 4-frame temporal block.
    }
    \label{fig:qual_grid_ucf}
    \vspace{0.2in}
\end{figure*}
\begin{figure*}[]
% \vspace{-0.1in}
    \centering
    \includegraphics[width=\linewidth]{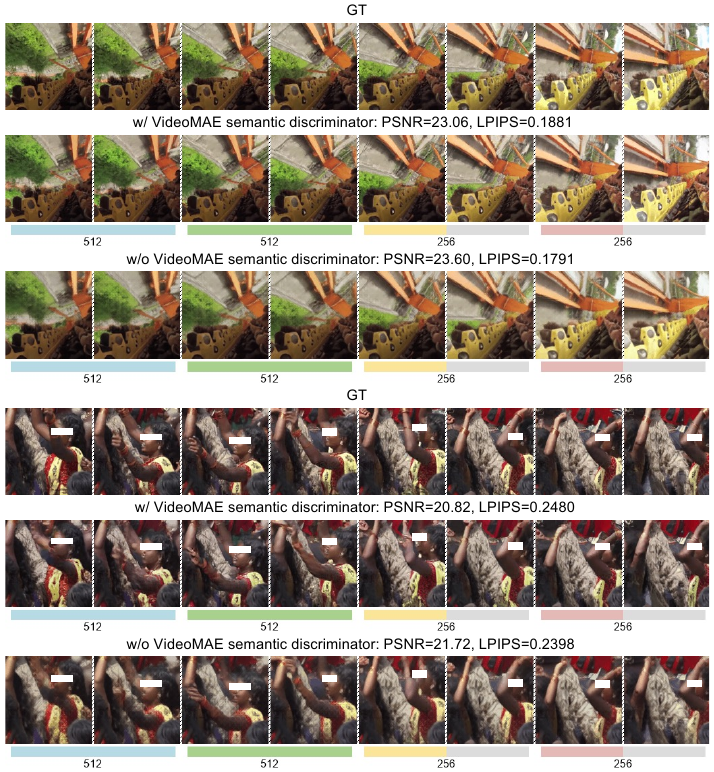}
    % \vspace{-0.30in}
    \caption{\textbf{Qualitative comparison for using VideoMAE discriminator or not.} Using VideoMAE discriminator can degrade the PSNR/LPIPS, but in actual perceptual checking, we find that this degradation is traded for alleviated blurriness and artifact patterns. Zoom in to check the visual details.
    }
    \label{fig:qual_grid_video_sem_disc}
    \vspace{0.4in}

\end{figure*}
\begin{figure*}[]
% \vspace{-0.1in}
    \centering
    \includegraphics[width=\linewidth]{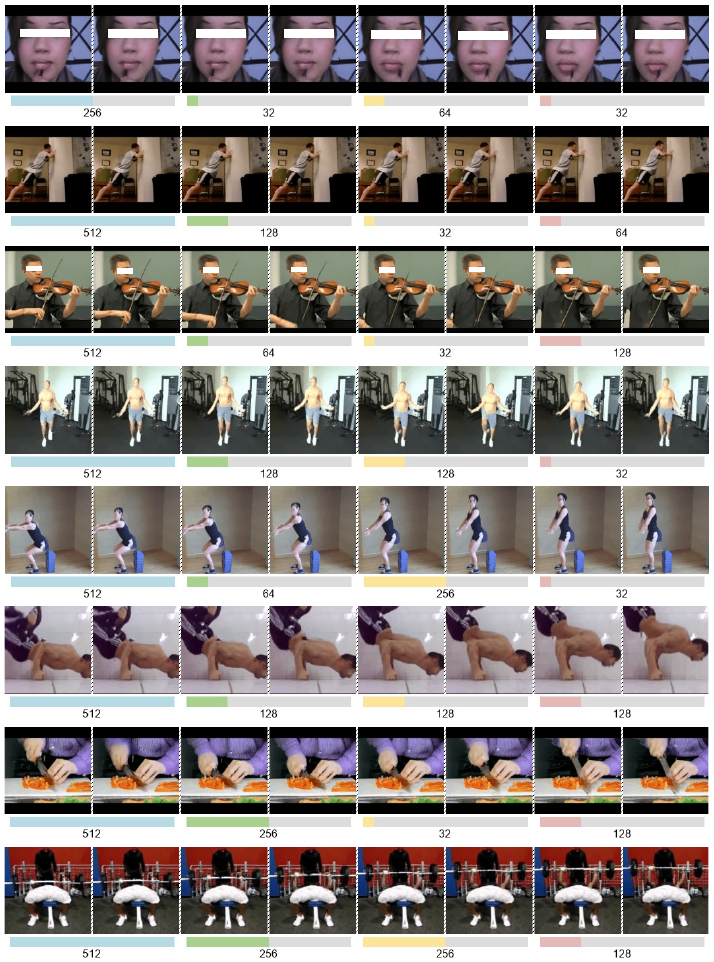}
    % \vspace{-0.30in}
    \caption{\textbf{Adaptive generation results on UCF-101.} We use the 633M GPT model trained on \ours{}. We use a constant 3.0 CFG for sampling.
    }
    \label{fig:qual_grid_ucf_gen}
\end{figure*}
\begin{figure*}[]
% \vspace{-0.1in}
    \centering
    \includegraphics[width=\linewidth]{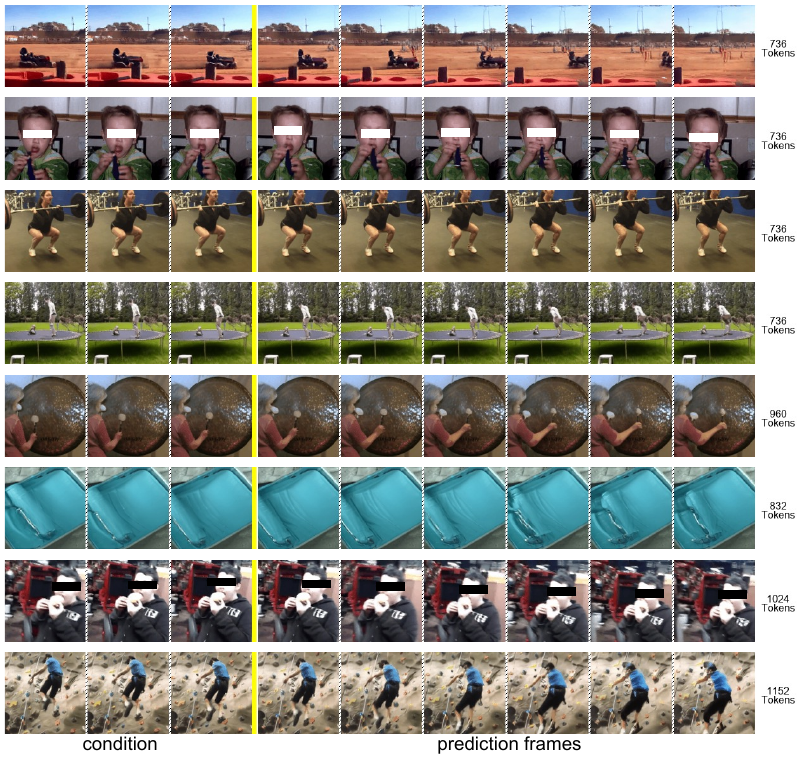}
    % \vspace{-0.30in}
    \caption{\textbf{Adaptive generation results on K600 frame prediction.} We use the 633M GPT model trained on \ours{}. We don't use CFG for sampling, following typical approaches. The 1, 3, 5 frames from the 5 condition frames are plotted as the condition part, and the rest 11 frames are downsampled into 6 frames for visualization.
    }
    \label{fig:qual_grid_k600_fp}
\end{figure*}

In this section, we present the full metrics of our final adaptive tokenizers on reconstruction in Tab.~\ref{tab:reconstruction_final}, and their reconstruction examples on samples from WebVid and the UCF-101 dataset. We also qualitatively present the effect of using VideoMAE~\cite{tong2022videomae} as part of the video discriminator. Besides, for video generation, we present generated samples on the UCF-101 class-to-video and K600 frame prediction task. 

\subsection{Adaptive Length Reconstruction Examples} 
We present the reconstruction results of our final video adaptive tokenizer, along with their token assignments decided by the router. In Fig.~\ref{fig:qual_grid_webvid}, we present the reconstruction results of the final adaptive tokenizer trained on WebVid-10M~\cite{Bain21WebVid}. And in Fig.~\ref{fig:qual_grid_ucf} are the reconstruction results on UCF-101 dataset, using another final video adaptive tokenizer trained on UCF-101 and K600 datasets. The patterns of video and assignment pairs given by the router correspond to intuitions. The router typically assigns more tokens to the initial temporal block, which helps the reconstruction of subsequent frames to also benefit from more precise information encoded for the initial block. Content with simple layouts or largely repeats previous frames receives fewer tokens. In contrast, videos that vary intensely are assigned more.

\subsection{VideoMAE Discriminator for Visual Quality}

In our ablation study in the main paper, the application of the VideoMAE~\cite{tong2022videomae, wang2022internvideo} discriminator significantly improves the rFVD and downstream gFVD but leads to degradation in PSNR and LPIPS. In this part, we aim to qualitatively examine the perceptual effect for improved rFVD and degraded  PSNR/LPIPS. We compare two video adaptive length tokenizers on WebVid-10M, one is trained with the pretrained VideoMAE discriminator, while another is trained with the PatchGAN discriminator. As shown in Fig.~\ref{fig:qual_grid_video_sem_disc}, although the reconstructed videos of the tokenizer trained with VideoMAE discriminator show worse PSNR and LPIPS, they are actually more perceptually preferable as they largely alleviate the blurriness or artifact patterns, especially for highly dynamic and challenging examples. Therefore, we conclude that despite the degradation in PNSR/LPIPS, VideoMAE discriminator still largely enhances the reconstruction quality perceptually.

\subsection{Adaptive Length Video Generation Examples}

We present UCF-101 class-to-video generation results in Fig.~\ref{fig:qual_grid_ucf_gen} and K600 frame prediction results in Fig.~\ref{fig:qual_grid_k600_fp}. As in Fig.~\ref{fig:qual_grid_ucf_gen}, the AR generation model learns an intuitive way for adaptive length generation. First, the model tends to pay more efforts for the generation of the first temporal block, which lays the foundation for the later generation. For later blocks, content with more variation tends to take more tokens to generate, while small-motion content takes fewer tokens.

\section{Computational Overhead Analysis}
\label{sec:compute_analysis}

\begin{table*}[ht]
\centering
\small
\begin{tabular}{llccccc}
\toprule
\textbf{Stage / Task} & \textbf{Model Size} & \textbf{Dataset} & \textbf{Bsz} & \textbf{Iters / Epochs} & \textbf{GPUs} & \textbf{Time} \\
\midrule
Stage 1: Proxy tokenizer training & 145M & WebVid~(or UCF\& K600) & 128 & 400k iters & 64×V100 & 116 h \\
Stage 2: Data curation & -- & WebVid 100k-Subset & -- & -- & 4×64×V100 & 12.5 h \\
Stage 3: Router training & 20M & WebVid 100k-Subset & 128 & 50k iters & 32×V100 & 5 h \\
Stage 4: Final adaptive tokenizer training & 145M & WebVid~(or UCF\& K600) & 128 & 1000k iters & 64×V100 & 347 h \\
\midrule
AR training: UCF-101 class-to-video & 633M & UCF-101 & 128 & 3000 epochs & 64×V100 & 88 h \\
AR training: K600 frame prediction & 633M & Kinetics-600 & 128 & 75 epochs & 64×V100 & 140 h \\
\bottomrule
\end{tabular}
\caption{Summary of compute and time for the four-stage tokenizer pipeline and subsequent AR model training.}
\label{tab:compute analysis}
\end{table*}

% Proxy tokenizer training
% Data curation analysis
% router training analysis
% Video tokenizer training
% Video AR model training
% AR inference

We present the computation cost for the model training of our four-stage framework and the downstream AR models, as shown in Tab.~\ref{tab:compute analysis}. Compared to the previous fixed-length methods, the extra training cost of our four-stage framework comes from the first three stages. However, the first three stages only take around 27.8\% of the total four-stage training cost. This percentage can be further reduced in real-world applications. The size of the proxy tokenizer and its training duration can be decreased for faster training, as we only need the proxy tokenizer to compare assignments, instead of performing well. The data curation can be processed by parallel and independent processes, without any GPU communication bottlenecks. And ultimately, the extra cost in adaptive tokenizer training is a one-time investment, but the savings for downstream deployment will consistently take effect. Therefore, the extra cost of the four-stage training is controllable and worthwhile.

\section{Attention Mask for \ours{}}
\label{sec:attn mask}
\begin{figure*}[t]
% \vspace{-0.1in}
    \centering
    \includegraphics[width=0.8\linewidth]{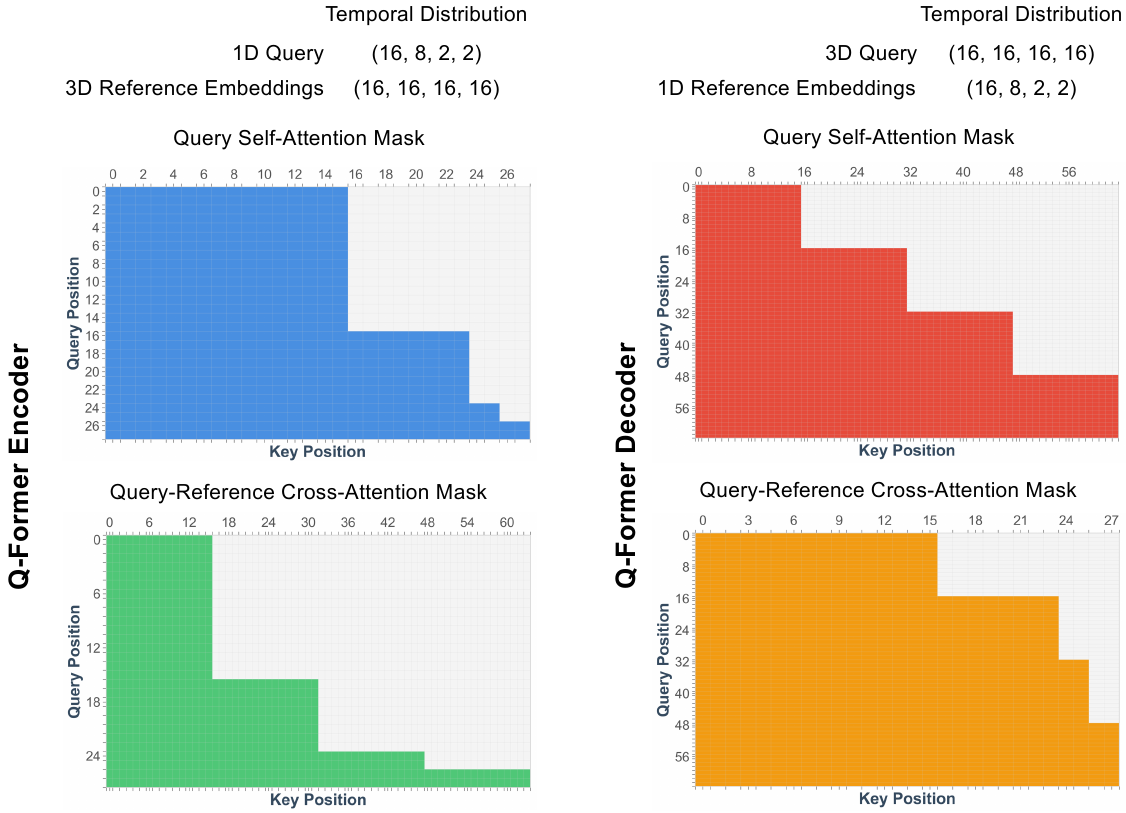}
    % \vspace{-0.30in}
    \caption{\textbf{Example for the attention masks in our Q-Former style adaptive tokenizer.} 
    }
    \label{fig:attn mask}
    \vspace{-0.10in}
\end{figure*}

In this section, we illustrate the specific details of the attention mask mechanism in our Q-Former style tokenizer, which ensures the temporal causal structure of our 1D token sequences. Each Q-Former layer consists of one self-attention module and one cross-attention module. The queries are first passed through the self-attention module, and then in the cross-attention module, the queries will attend to the reference embeddings. Next, we use an example to show what attention masks look like in the Q-Former encoder and Q-Former decoder. Assume a video clip is patchified into a $4\times 4\times4$ shape tensor, where the first $4$ corresponds to the number of temporal blocks. And let the assignment of tokens across the $4$ blocks for this video be $(16,8,2,2)$. Then, the attention masks for the Q-Former encoder and Q-Former decoder will be the ones shown in Fig.~\ref{fig:attn mask}. The query embeddings of each temporal block can only attend to query embeddings or reference embeddings that are no later than this temporal block.

\begin{table}[]
    \centering
    \resizebox{0.5\textwidth}{!}{
        \begin{tabular}{llcc}
        \toprule
        \textbf{Dataset} & \textbf{Method} & \textbf{Val top1/top5 acc.} & \textbf{Proxy Reward Percentile} \\
        \midrule
        \multirow{2}{*}{WebVid} & Best-Uniform & - & 84.88\% \\
        & Router & 11.72\% / 35.03\% & 96.96\% \\
         \midrule
        \multirow{2}{*}{UCf-101}  & Best-Uniform & - & 88.46\% \\
        & Router & 5.77\% / 23.68\% & 96.19\% \\

        \bottomrule
        \end{tabular}
    }
    \caption{\textbf{Accuracy \vs proxy reward percentile for the router assignment.} In terms of accuracy. the assignments predicted by the router do not usually hit the top1 or top5 highest proxy reward assignments. However, in terms of proxy reward percentile, the router assignments achieve good results, and generalize to unseen dataset~(UCF-101) well. 
    }
    \label{tab:router_acc_reward}

\end{table}

\section{Accuracy \vs Proxy Reward for Routers}
\label{sec:router_analysis}

% In this part, we examine the accuracy of the router on the validation sets. We find that although the accuracy is relatively slow, the assignments given by the router still obtain decent proxy reward. We use preference weights $w_q=1.2, w_l=0.8$ for proxy reward calculation, which are the same as the weights used for the evaluated router training data curation. To evaluate relatively how good the predicted assignments are among all candidate assignments, we use a new metric: proxy reward percentile. 

% As shown in Tab.~\ref{tab:router_acc_reward},  

% \tianwei{Consider moving to supp.}

In this part, we examine the accuracy of the router on the validation sets. We find that although the accuracy is relatively slow, the assignments given by the router still obtain decent proxy reward. We use preference weights $w_q=1.2, w_l=0.8$ for proxy reward calculation, which are the same as the weights used for the evaluated router training data curation. To evaluate relatively how good the predicted assignments are among all candidate assignments, we use a new metric, proxy reward percentile, defined as:
\begin{equation}
    \mathcal{P} = 
    \frac{\mathbb{E}_{x}(R_\text{proxy}(a_\text{eval}, x)) - \mathbb{E}_{x}(R_\text{proxy}(a_\text{worst}, x))}
    {\mathbb{E}_{x}(R_\text{proxy}(a_\text{best}, x)) -
    \mathbb{E}_{x}(R_\text{proxy}(a_\text{worst}, x))
    } \times 100\%
\end{equation}
where $a_\text{eval}$ is the assignment to be evaluated for video $x$, $a_\text{best}$ is the searched max-proxy-reward assignment for $x$, and $a_\text{worst}$ is the min-proxy-reward assignment. In practice, $a_\text{eval}$ can be given by the router according to $x$ or by some other manually designed strategy. $\mathbb{E}_{x}(R_\text{proxy}(a_\text{eval}, x))$ is the expectation for the proxy reward based on $a_\text{eval}$ and $x$. The range of $\mathcal{P}$ is $[0,1]$ and the larger $\mathcal{P}$, the better the assignment strategy is. We design a best-uniform baseline, which chooses the max-proxy-reward uniform assignment for $x$, to compare with the router assignment. As shown in Tab.~\ref{tab:router_acc_reward}, on WebVid validation set, the top1 accuracy of the router is relatively low, but the proxy reward percentile of the router is high. Moreover, when tested on the unseen UCF-101 dataset, although the top1 accuracy of the router significantly drops, its proxy reward percentile is largely maintained. This phenomenon indicates that the router does not need to be very precise to achieve good performance, implying that the optimal assignment prediction task is not demanding, and some deviation from the best choice won't result in a large performance drop.

\begin{figure}[]
% \vspace{-0.1in}
    \centering
    \includegraphics[width=\linewidth]{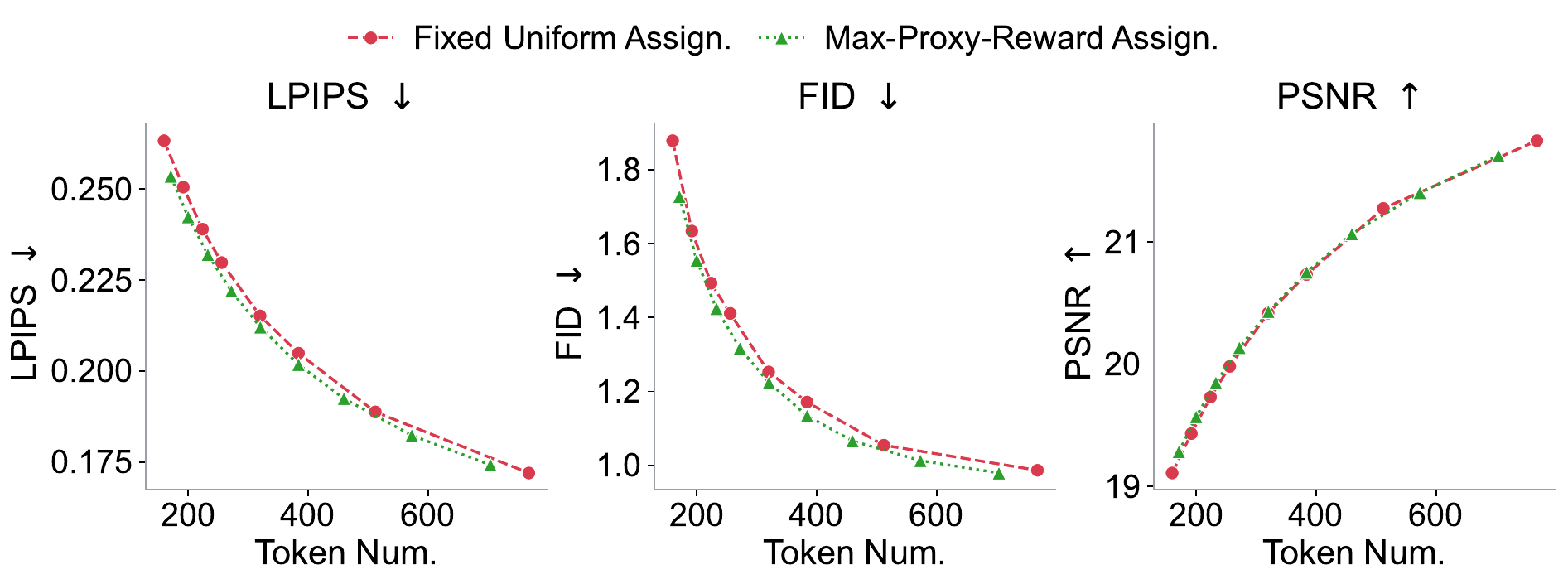}
    % \vspace{-0.30in}
    \caption{\textbf{Image tokenization quality-cost trade-off curve.} On ImageNet $256\times 256$ reconstruction, the improvements of max-proxy-reward assignment can be marginal compared to uniform assignment.
    } 
    \label{fig:trade_off_curve_img}
\end{figure}
\begin{table}[]
    \centering
    \resizebox{0.5\textwidth}{!}{
        \begin{tabular}{lccccc}
        \toprule
         & \textbf{LPIPS$\downarrow$} & \textbf{rFID$\downarrow$} & \textbf{\#rTokens$\downarrow$}&  \textbf{gFID$\downarrow$} &
         \textbf{\#gTokens$\downarrow$}
         \\
        \midrule
        Uniform~(Final)      & 0.2205 & \textbf{1.22}     & 256   &   4.72 & 256\\
         Router~(Final)  & 0.2455 & 1.46    & 205~\small{(-19.9\%)}    &  \textbf{4.51} & 197~\small{(-23.0\%)} \\
        \bottomrule
        \end{tabular}
    }
    \caption{\textbf{Image final tokenizer validation.} For ImageNet $256\times 256$, saving 19.9\% tokens by adaptive tokenization inevitably leads to worse rFID, but the performance and efficiency of downstream AR generation models can still benefit from our router. The AR generation models use a constant 1.5 CFG during inference.
    }
    \label{tab:final_validation_img}

\end{table}
\section{Image Adaptive Tokenization}
\label{sec:img adaptive exp}

Different from videos, images don't have a temporal dimension, so intuitively images are much less redundant than videos. Our experiments on ImageNet~\cite{russakovsky2015imagenet} $256\times 256$ show that in this setting, the improvement in overall reconstruction quality that can be brought by assigning different token lengths to different images could be limited. However, for downstream generation, adaptive image tokenization can still help produce better generation FID~\cite{FID} with fewer tokens generated. This result highlights that training generative models on adaptive length sequences is not only efficient but also beneficial to their generation capability.

\subsection{Implementation Details}
We train image tokenizers on $256\times 256$ ImageNet~\cite{russakovsky2015imagenet} dataset using a similar CNN + Q-Former hybrid architecture of GigaTok-S-B~\cite{Xiong2025GigaTok}. The basic training recipe is also largely aligned to GigaTok except that we utilize DINOv3~\cite{simeoni2025dinov3} to provide semantic alignment. We train all image tokenizers with 256 batch size with only 400k iterations, as we only target to validate the gain of adaptive tokenization on images compared to the fixed-length baseline. Our four-stage framework smoothly translates from videos to image adaptive tokenization, because an image can be equivalent to a one-block video in the tokenization process.

For the proxy tokenizer, we predefine 8 candidate levels of token numbers, $\{512, 384, 256, 192, 128, 96, 64, 32\}$, to be assigned to each image for variable length tokenizer training. For image router training, we train ViT-S~\cite{vit} size routers on a subset of ImageNet training split of 100k images. They are trained for 25k iterations with a batch size of 256. We use normalized LPIPS as the quality metric in the proxy reward calculation. The reconstruction quality is evaluated on the 50k ImageNet validation set. For downstream AR generation validation, we train Llama-like~\cite{llamagen} GPT-B models on each tokenizer for 300 epochs on ImageNet, and evaluate them with generation FID~\cite{FID} with a constant 1.5 CFG, following the typical approaches~\cite{llamagen, adm}.

\subsection{Results}
\noindent\textbf{Quality-cost trade-off curve.} We use a similar way as for video proxy tokenizers, to plot the quality-cost trade-off curve under different overall token budgets on an image proxy tokenizer. As shown in Fig.~\ref{fig:trade_off_curve_img}, the quality-cost trade-off curve evaluated on image proxy-tokenizers shows that the improvements brought by max-proxy-reward assignment are limited, which is different from the results on videos. This phenomenon corresponds to observations in previous adaptive image tokenization trials~\cite{Shen2025CAT, mao2025dove} on ImageNet, where their adaptive length image tokenizers cannot outperform their fixed-length baselines even with the same overall token budgets.

\noindent\textbf{Final image tokenizer validation.} In the final image adaptive tokenizer training, as shown in Tab.~\ref{tab:final_validation_img}, we utilize an image router trained with $w_q=1.3, w_l=0.7$ to save 19.8\% tokens in reconstruction, but it inevitably leads to worse rFID. However, the AR model trained on our adaptive image tokenizer achieves better gFID with 23.0\% fewer tokens generated, compared to the fixed-uniform baselines, which assign 256 tokens to all $256\times256$ images. This indicates that the performance and efficiency of downstream AR image generation can still benefit from image adaptive tokenization using our method.

\end{document}